\documentclass[lettersize,journal,biography]{IEEEtran}  
\usepackage{amsmath,amsfonts}
\usepackage{array}
\usepackage[caption=false,font=normalsize,labelfont=sf,textfont=sf]{subfig}
\usepackage{textcomp}
\usepackage{stfloats}
\usepackage{url}
\usepackage{verbatim}
\usepackage{cite}
\usepackage{multicol}
\usepackage{amsmath,amssymb,amsfonts}
\usepackage{textcomp}
\usepackage{cite}
\usepackage{xcolor}
\usepackage{mathrsfs}
\usepackage{lipsum} 
\usepackage{balance} 
\usepackage{amsfonts}
\usepackage{cite}

\usepackage{indentfirst}
\usepackage[ruled,vlined]{algorithm2e}
\usepackage{flushend}
\usepackage{colortbl}
\usepackage{graphicx}  
\usepackage{booktabs}
\usepackage{threeparttable}
\usepackage{hyperref}
\usepackage{multirow}
\usepackage{float}
\usepackage{nomencl}
\usepackage{etoolbox}
\def\BibTeX{{\rm B\kern-.05em{\sc i\kern-.025em b}\kern-.08em
    T\kern-.1667em\lower.7ex\hbox{E}\kern-.125emX}}
\usepackage{balance}

\begin{document}

\title{GSM-GS: Geometry-Constrained Single and Multi-view Gaussian Splatting for Surface Reconstruction}
\author{Xiao Ren, Yu Liu, Ning An, Jian Cheng, Xin Qiao, and He Kong
\thanks{Xiao Ren, Yu Liu, and He Kong (corresponding author) are with the School of Automation and Intelligent Manufacturing, Southern University Science and Technology, Shenzhen 518055, China; Emails: [{12431359,12250044}]@mail.sustech.edu.cn; kongh@sustech.edu.cn. Ning An and Jian Cheng are with the Research Institute of Mine Artificial Intelligence, China Coal Research Institute, and the State Key Laboratory of Intelligent Coal Mining and Strata Control, Beijing 100031, China; Emails: ning.an.010@foxmail.com; jiancheng@tsinghua.org.cn. Xin Qiao is with the Institute of Artificial Intelligence and Robotics, Xi’an Jiaotong University, 710049, Shaanxi, China; Email: wudiqx@xjtu.edu.cn.}}

\markboth{}
{GSM-GS: Geometry-Constrained Single and Multi-view Gaussian Splatting for Surface Reconstruction}

\maketitle

\begin{abstract}
Recently, 3D Gaussian Splatting has emerged as a prominent research direction owing to its ultrarapid training speed and high-fidelity rendering capabilities.
However, the unstructured and irregular nature of Gaussian point clouds poses challenges to reconstruction accuracy.
This limitation frequently causes high-frequency detail loss in complex surface microstructures when relying solely on routine strategies.
To address this limitation, we propose GSM-GS: a synergistic optimization framework integrating single-view adaptive sub-region weighting constraints and multi-view spatial structure refinement.
For single-view optimization, we leverage image gradient features to partition scenes into texture-rich and texture-less sub-regions. 
The reconstruction quality is enhanced through adaptive filtering mechanisms guided by depth discrepancy features. 
This preserves high-weight regions while implementing a dual-branch constraint strategy tailored to regional texture variations, thereby improving geometric detail characterization. 
For multi-view optimization, we introduce a geometry-guided cross-view point cloud association method combined with a dynamic weight sampling strategy. 
This constructs 3D structural normal constraints across adjacent point cloud frames, effectively reinforcing multi-view consistency and reconstruction fidelity.
Extensive experiments on public datasets demonstrate that our method achieves both competitive rendering quality and geometric reconstruction.
See our interactive \href{https://aislab-sustech.github.io/GSM-GS/}{\textcolor{blue}{project page}}.
\end{abstract}


\begin{figure*}[h!]
    \centering
    \includegraphics[width=\textwidth]{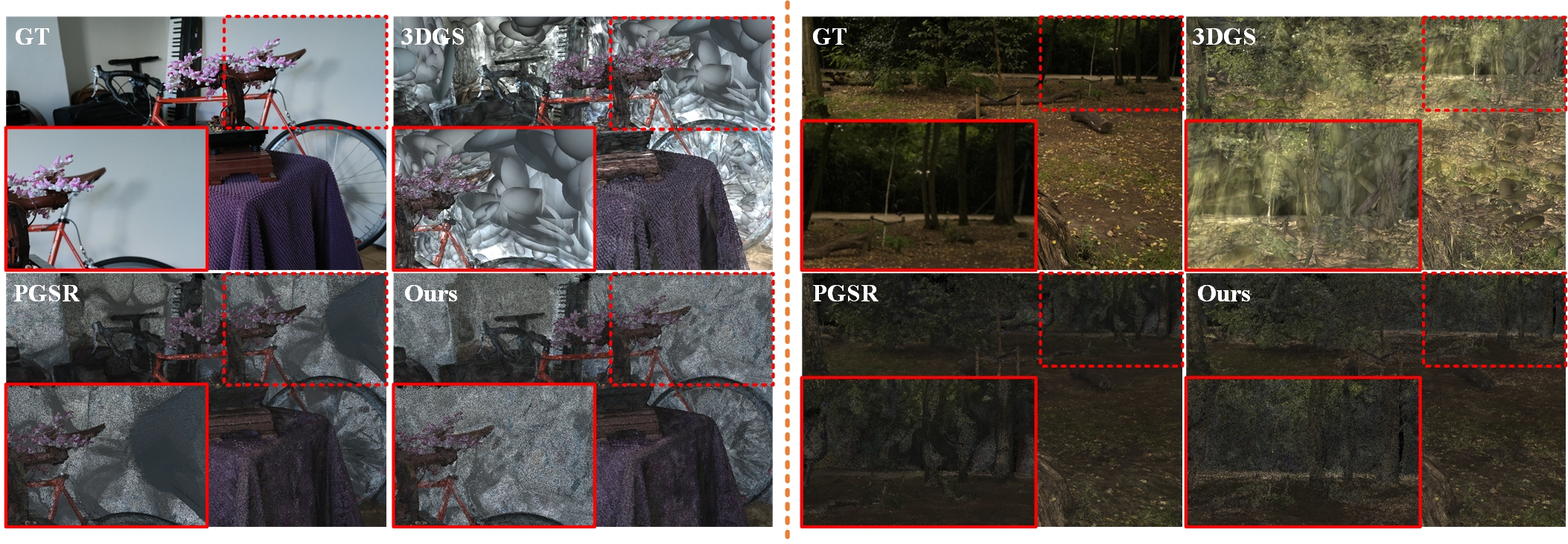}
    \caption{Spatial distribution comparison of Gaussian ellipsoids. This figure compares the spatial ellipsoid distributions reconstructed by the 3DGS, PGSR, and Ours algorithms for the rendered scene. The 3DGS results exhibit suboptimal performance, as their Gaussian ellipsoids fail to conform closely to object surfaces. While both PGSR and ours employ thin Gaussian ellipsoids for surface approximation, our method introduces novel constraints derived from single-view and multi-view paradigms. This optimization yields more regular ellipsoid distributions and significantly enhanced surface conformity.}
    \label{fig:first_pic}
\end{figure*}

\section{Introduction}
New View Synthesis (NVS) and geometric surface reconstruction have broad applications spanning robotic environmental perception~\cite{SG-SLAM, su2021necessary, wakulicz2021active, LVI_GS} and navigation~\cite{Li_Navigation_2024, Xu_Navigation_2023, Wang_navigation_2021, SLAM_Based}, AR/VR systems~\cite{deng2022fov}, 3D content generation/editing~\cite{yi2024gaussiandreamer}, and autonomous driving~\cite{wang2024dc, yang2025drgs}. 
Conventional multi-view stereo (MVS) methods typically estimate depth information through cross-view feature matching, followed by depth fusion to reconstruct object surfaces. 
While these methods achieve satisfactory performance in many scenarios, they demand significant computational resources and struggle to accurately capture surface texture details in complex scenes while ensuring multi-view consistency.

Among existing NVS and 3D reconstruction-related methods, Neural Radiance Field (NeRF)~\cite{mildenhall2021nerf} implicitly represents the 3D scene and achieves new view rendering through an analysis-by-synthesis approach. 
As discussed in the existing literature~\cite{yariv2021volume}, due to constraints imposed by implicit representations, the original NeRF method typically produces 3D reconstructions with limited accuracy, and its ray-sampling and volume-rendering paradigm requires training times of tens of hours. 
Although recent works, including Instant-NGP~\cite{muller2022instant}, reduce the training time to a few minutes, the reconstruction performance still needs considerable improvement.

Recently, the 3D Gaussian Splatting (3DGS)~\cite{kerbl20233d} technique has been widely used in the field of 3D reconstruction and new view synthesis due to its high-quality rendering and ultrashort training time.
The technique utilizes a set of anisotropic Gaussian ellipsoids to represent the 3D scene, where each Gaussian ellipsoid contains physical attributes such as position, color, opacity, and covariance parameters.
Despite the advantages of the 3DGS method, it still faces challenges in high-fidelity reconstruction~\cite{huang2024textured} and rendering~\cite{shen2025topology}. 

For example, in surface texture-rich regions, the optimization process based on photometric consistency constraints is susceptible to interference from high-frequency texture components, leading to ambiguous decoupling between normal vector orientation and Gaussian ellipsoid anisotropy covariance parameters.  
Moreover, in texture-less or uniformly shaded regions, the lack of sufficient texture gradient information makes normal vector estimation prone to local minima, resulting in over-smoothed geometric details and distorted surface curvature.
Across multiple views, the absence of cross-view geometry-appearance constraints for discrete Gaussian primitives leads to reconstruction artifacts and geometric distortion~\cite{yu2024gsdf} in novel view synthesis.

In this paper, we propose a novel surface reconstruction and novel view rendering framework that improves reconstruction and rendering quality through single-view and multi-view geometry-guided constraints.
On the one hand, for single-view optimization, we first achieve adaptive partitioning of texture-rich and texture-less regions based on image gradient features and construct confidence factors using depth discrepancy metrics to dynamically filter high-confidence regions. 
Differential normal vector constraints are then adaptively applied according to regional characteristics to enhance fine surface detail capture, which effectively suppresses high-frequency geometric feature degradation during iterative optimization. 

On the other hand, to improve multi-view consistency, we utilize adjacent point cloud frames across views and simultaneously compute weight information derived from depth discrepancy transformations in neighboring viewpoints.
These weights are fused to generate global confidence metrics, which sample stable geometric priors from paired point cloud frames to establish spatial normal consistency constraints. 
By integrating confidence-guided hierarchical geometric constraints across single and multiple views, we build continuous geometric correlation models between unstructured point clouds.
This approach overcomes inter-view structural misalignment caused by discrete Gaussian distributions while enhancing surface detail reconstruction accuracy.
The spatial distribution of the Gaussian ellipsoid before and after optimization is shown in Fig.~\ref{fig:first_pic}. Our contributions can be summarised as follows:
\begin{itemize}
    \item We propose a single-view adaptive sub-region constraint with dual-branch optimization, which guides scene decoupling and weight filtering through image gradients for accurate reconstruction of high-frequency geometric features such as surface microstructures.
    \item We introduce a weight-guided dynamic sampling strategy with cross-view geometric normal correlation constraints, constructing a 3D continuous geometric correlation model via multi-view global weight adaptive filtering of point cloud data, effectively addressing multi-view consistency loss in complex scenes.
    \item A comprehensive algorithmic evaluation is conducted on benchmark datasets (DTU~\cite{jensen2014large}, Mip-NeRF360~\cite{barron2022mip}, and Tanks and Temples~\cite{knapitsch2017tanks}), including systematic comparative evaluation with advanced 3D reconstruction frameworks and detailed analysis of surface reconstruction and novel view synthesis performance.
\end{itemize}

The remainder of this paper is organized as follows. Section~\ref{app:related_work} reviews related work on 3D reconstruction and Gaussian Splatting. Section~\ref{app:preliminary_plannar_based} introduces the preliminaries of Planar-based 3D Gaussian Splatting. Section~\ref{app:methodology} details our proposed GSM-GS framework. Section~\ref{app:experiements} presents the experimental implementation, comparative results, and ablation studies. Finally, Section~\ref{app:conclusion} concludes the paper.

\section{Related Work}
\label{app:related_work}
\subsection{Multi-view Stereo}
In 3D surface reconstruction, multi-view stereo is a classical problem.
MVS generates intermediate geometric representations such as point clouds, voxel meshes~\cite{fridovich2022plenoxels}, and depth maps through multi-view geometric matching.
Traditional methods like COLMAP~\cite{schoenberger2016sfm} rely on photometric consistency constraints in strongly textured regions and demonstrate excellent performance in structured scenes, but their performance degrades significantly in weakly textured or occluded regions. 
To address texture-deficient scenarios, learning-based MVS methods have been developed.
CasMVSNet~\cite{gu2020cascade, Depth_Restoration} employs cascaded cost volume filtering to enhance depth accuracy while reducing GPU memory consumption.
TransMVSNet~\cite{ding2022transmvsnet} integrates attention mechanisms from Transformer architectures to strengthen feature relevance. 
However, these methods could produce artifacts at occlusion boundaries and lack multi-view consistency due to their per-view depth prediction paradigm.

\subsection{Neural Radiance Fields}
The mapping of spatial coordinates to color and volume density in Neural Radiance Fields has driven significant advances in Novel View Synthesis. 
For instance, Mip-NeRF~\cite{barron2021mip} employs conical frustums to characterize scenes at multiple scales, while Mip-NeRF 360~\cite{barron2022mip} extends this framework to handle unbounded scenes.
BungeeNeRF~\cite{xiangli2022bungeenerf} adopts a progressive residual learning strategy to capture large-scale scene details hierarchically. 
Pyramid NeRF~\cite{zhu2023pyramid} proposes a coarse-to-fine reconstruction approach by progressively augmenting high-frequency details using image pyramids. 
BARF~\cite{lin2021barf} introduces frequency-scheduled positional encoding to enable training without precise camera poses. 
Another critical direction in NeRF research is high-fidelity 3D scene reconstruction.
Methods like NeuS~\cite{wang2021neus} and BakedSDF~\cite{yariv2023bakedsdf} leverage signed distance functions to represent surfaces and ensure watertight geometry reconstruction.
Neuralangelo~\cite{li2023neuralangelo} combines multi-resolution hash grids with SDFs for large-scale scene modeling, while Nerf2Mesh~\cite{zhang2024voxel} implements reprojection-error-driven adaptive optimization to co-optimize mesh vertex distributions and volumetric density parameters.
Although NeRF-based frameworks achieve impressive surface reconstruction and rendering quality, balancing training efficiency with reconstruction fidelity remains a challenge.

\subsection{3D Gaussian Splatting}
3D Gaussian Splatting has become a prominent method for scene rendering and surface reconstruction. 
Its explicit representation using anisotropic 3D Gaussian primitives enables rapid reconstruction and real-time rendering~\cite{li2024geogaussian}. 
Current optimized variants of 3DGS address rendering, dynamic scenes, and large-scale reconstruction~\cite{gao2024cosurfgs,lin2024vastgaussian}. 
For instance, DNGaussian~\cite{li2024dngaussian} employs depth regularisation to achieve high-quality novel view synthesis with reduced sampling.
DN-Splatter~\cite{turkulainen2025dn} enhances Gaussian primitive representations through base model optimization. 
To improve geometric fidelity, RaDe-GS~\cite{zhang2024rade} introduces depth/normal map rendering during rasterization.
GOF~\cite{yu2024gaussian} leverages Gaussian opacity fields for surface reconstruction, and SuGaR~\cite{guedon2024sugar} applies regularisation constraints to align Gaussians with surfaces, extracting Poisson-reconstructed meshes via density fields. 
2DGS~\cite{huang20242d} constrains Gaussian primitives by setting their $z$-axis covariance to zero for planar projection. 
SolidGS~\cite{shen2024solidgs} integrates Gaussian surfel splatting to optimize primitive representations. 
PGSR~\cite{chen2024pgsr} incorporates unbiased depth estimation and multi-view consistency constraints for accurate surface reconstruction. 
MPGS~\cite{MPGS} constrains Gaussian ellipsoids onto multiple planes to reduce redundancy based on planar priors, enhancing operational speed while preserving rendering quality.
PUP 3D-GS~\cite{PUP_3D-GS} introduces an uncertainty-based pruning method to quantify Gaussian importance, reducing computational overhead while improving reconstruction accuracy.
Furthermore, SpecTRe-GS~\cite{SpecTRe_GS} integrates a ray-tracing mechanism within the 3D space to achieve high-fidelity rendering of specular surfaces.

Although existing methods have demonstrated notable improvements in reconstruction quality, their performance remains to be enhanced for texture-less scenarios. To address this, we introduce specialized constraints for texture-less regions and refine the 3D point cloud reconstruction, thereby improving reconstruction fidelity.
Different from mesh-integrated approaches like GaMeS~\cite{gao2024mesh} that trade precision for editing flexibility via fixed topological priors, our method prioritizes high-fidelity reconstruction via depth and normal consistency.

\begin{figure*}[h]
    \centering
    \includegraphics[width=\textwidth]{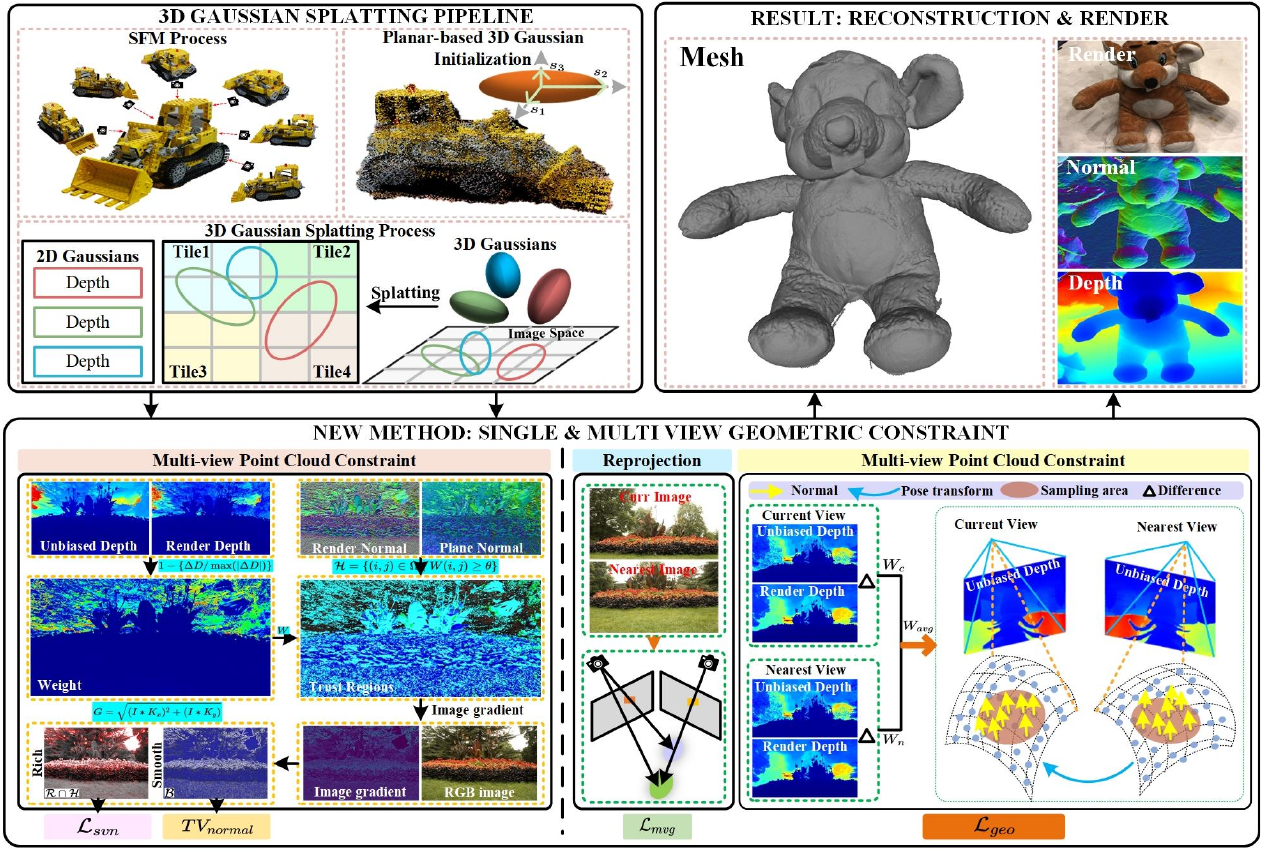}
    \caption{GSM-GS Overview. The algorithm framework takes sparse point cloud data and image input, initializes each point cloud into a thin Gaussian ellipsoid, and processes a single-view adaptive partitioning constraint, dual-branch optimization strategy, weight-guided dynamic sampling strategy, and cross-view geometric correlation normal constraint. The system is innovatively optimized from the perspectives of single-view and multi-view, and finally outputs high-quality reconstruction and rendering results.}
    \label{fig:systemFramework}
\end{figure*}

\section{Preliminary: Planar-based 3D Gaussian Splatting}
\label{app:preliminary_plannar_based}
The 3D Gaussian Splatting framework employs a collection of anisotropic ellipsoids ${\mathcal{G}_i}$ to represent physical scenes, where each ellipsoid follows a Gaussian distribution parameterized by its centroid position $\mu_i \in \mathbb{R}^3$, opacity $\alpha_i$, spherical harmonic (SH) coefficients, and covariance matrix $\Sigma_i \in \mathbb{R}^{3\times3}$. These parameters collectively determine the ellipsoid's color $c_i \in \mathbb{R}^3$ and geometric shape. 
Formally, the Gaussian ellipsoid distribution in the world coordinate system is defined as:
\begin{equation}
    \mathcal{G}_{i}(x|\mu_{i},\Sigma_{i})=\exp \left\{-\frac{1}{2}(x - \mu_{i})^{\top}\Sigma_{i}^{-1}(x - \mu_{i})\right\},
\end{equation}
where the covariance matrix $\Sigma_{i}$ admits decomposition into a rotation matrix $R_{i} \in \mathrm{SO}(3)$ and scaling matrix $S_{i} \in \mathbb{R}^{3\times3}{\text{diag}}$, satisfying $\Sigma{i} = R_{i} S_{i} S_{i}^{\top} R_{i}^{\top}$. 
Through the transformation matrix $W \in \mathrm{SE}(3)$, the Gaussian ellipsoid is transformed from world to camera coordinates, yielding the projected mean vector 
$\mu'_i = K W [\mu_i^{\top}\ 1]^{\top}$ and covariance matrix $\Sigma'_i = J W \Sigma_i W^{\top} J^{\top}$. 
Here, $J \in \mathbb{R}^{2\times3}$ denotes the Jacobian matrix of perspective projection with radial distortion approximation, while $K \in \mathbb{R}^{3\times3}$ represents the camera intrinsic matrix.
Combining the color $c_i$ and opacity $\alpha_i$ of the Gaussian ellipsoid, the RGB image in the current viewpoint is rendered according to the $\alpha$ blend, and the specific process can be expressed as follows:
\begin{equation}
    C=\sum_{i \in P} T_{i} \alpha_{i} c_{i}, T_{i}=\prod_{j=1}^{i - 1}(1 - \alpha_{j}),
\end{equation}
where $P$ denotes the Gaussian ordered by depth and $T_i$ denotes the cumulative transmittance. 
Similarly, the normal map $N$ and the distance map $D$ of the scene can be rendered via $\alpha$-blending:
\begin{equation}
    N=\sum_{i \in |N|} T_{i} \alpha_{i} n_{i}, \text{ } D=\sum_{i \in |N|} T_{i} \alpha_{i} d_{i}.
\end{equation}
Here, $n_i$ denotes the normal vector of the Gaussian ellipsoid surface, and $d_i$ denotes the distance from the ellipsoid center to the camera center.
Based on the normal and depth maps, the unbiased depth map $\hat{D} \in \mathbb{R}^{H \times W}$ can be computed:
\begin{equation}
    \hat{D}(p) = \frac{D}{N(p) K^{-1} \tilde{p}},
\end{equation}
where $\tilde{p}$ denotes the homogeneous coordinate of the 2D image-plane position $p=[u, v]^\top$. 
In 3DGS, in addition to the photometric loss of the base image $\mathcal{L}_{rgb}$, in order to suppress the effect of floating points on the reconstruction quality, the homography matrix $H_{rn}$ is used to maintain the geometric multiview consistency $\mathcal{L}_{mvpro}$ and the photometry multi-view consistency $\mathcal{L}_{mvncc}$:
\begin{equation}
    \left\{\begin{array}{l} 
        \mathcal{L}_{mvpro}=\frac{1}{|\mathbb{V}|} \sum_{p_{r} \in \mathbb{V}}\left\| p_{r}-H_{nr}H_{rn} p_{r}\right) \| \\ \mathcal{L}_{mvncc}=\frac{1}{|\mathbb{V}|} \sum_{p_{r} \in \mathbb{V}}\left(1 - NCC\left(I_{r}\left(p_{r}\right), I_{n}\left(H_{rn} p_{r}\right)\right)\right) 
    \end{array}\right. ,
\end{equation}
where $p_r$ denotes the pixel position in the reference frame, $p_n$ is obtained by projecting $p_r$ to adjacent frames via the single response matrix $H_{rn}$, and $I_r(p_r)$ and $I_n(H_{rn}p_r)$ denote the pixel blocks of a particular size centred on $p_r$ and $p_n$; $\mathbb{V}={p_r|\left\|(p_{r}-H_{nr}H_{rn} p_{r})\right\| \le \theta}$ is the set of pixels for which the reprojection error has not exceeded a threshold; the normalized cross-correlation (NCC) value $NCC(I_r(p_r), I_n(p_n))$ measures the local patch similarity. The final multi-view geometric constraint is formulated as  $\mathcal{L}_{mvg}=\mathcal{L}_{mvpro}+\mathcal{L}_{mvncc}$.

\section{Methodology}
\label{app:methodology}
As shown in Fig.~\ref{fig:systemFramework}, the GSM-GS method enhances 3D Gaussian reconstruction accuracy through single and multi-view co-optimization. 
In this section, we detail the GSM-GS framework's methodology. 
In subsection~\ref{app:sub_region_adaptive}, we present the sub-region adaptive weighting strategy with single-view normal constraints; in subsection~\ref{app:geometry_guided}, we introduce cross-view geometric spatial normal constraints; in subsection~\ref{app:loss_functions}, we describe the loss formulation integrating reconstruction quality, rendering fidelity, and geometric accuracy optimization. Furthermore, to improve readability, definitions of mathematical symbols are described in detail in Appendix~\ref{app:symbol_definition}.
\begin{figure*}[h]
    \centering
    \includegraphics[width=\textwidth]{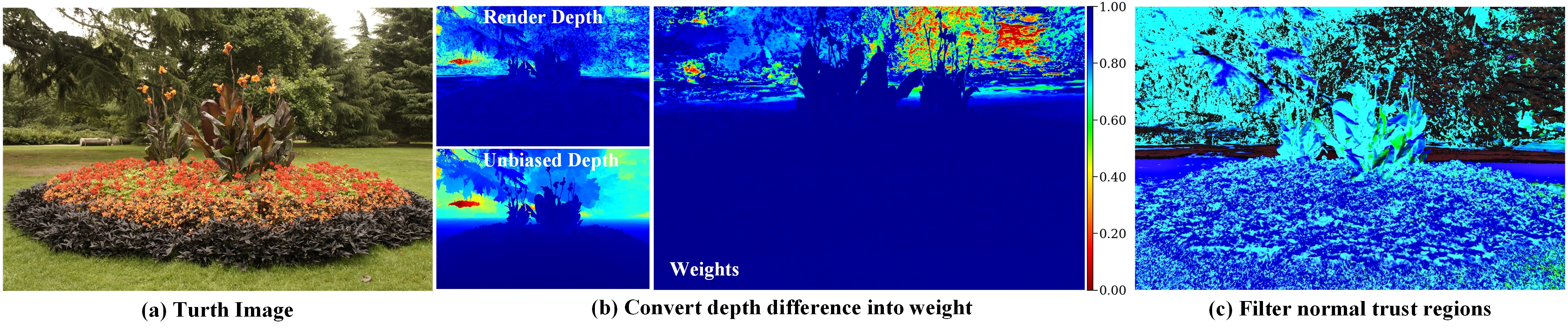}
    \caption{Filtering trust regions based on converting the difference between rendering depth and weight information. (a) is the RGB map of the real scene; (b) the difference between the rendering depth and the unbiased depth is used to calculate the difference, and then the difference is converted into weight information to reflect the reconstruction effect of each point in the scene; (c) the scene is partitioned using the weight information, and the low weight part is filtered out(the bright part of the map), and the low weight part is filtered out(the darker region of the map), and the data of the trust region is prioritized to added into the system for computation in the subsequent regularization. The data from the trusted regions is prioritized in the subsequent regularisation and added to the system for computation.}
    \label{fig:weightDisplay}
\end{figure*}
\subsection{Sub-region Adaptive Weighting with Single-view Normal Constraints}
\label{app:sub_region_adaptive}
In order to accurately characterize the surface features of the object in a single view, a sub-region optimization strategy is designed in this summary; the overall process is as in Algorithm~\ref{singleAlgorithm}.
First, image gradient analysis dynamically partitions the scene. 
High-confidence regions are then selected within partitioned areas using weight metrics.
To adapt to regional characteristics and refine error propagation in the optimization process, we implement differentiated error computation methods across regions, augmented by confidence-aware weighting to modulate error impact.

\subsubsection{Reliance on Regional Screening}
This framework employs two depth computation methods from~\cite{chen2024pgsr} to obtain the rendered depth map $D$ and unbiased depth map $\hat{D}$.
When the reconstruction and rendering quality approaches photorealistic accuracy, the depth discrepancy between $D$ and $\hat{D}$ diminishes.
The depth representations, therefore, satisfy the asymptotic consistency condition: $\lim | D-\hat {D}| \to 0$ as geometric reconstruction converges to the true surface. This discrepancy is consequently transformed into a weight metric $W \in [0,1]$ through:
\begin{equation}
    W(i, j) = 1 - \frac{\Delta D(i, j)}{\|\Delta D\|_{\infty, \Omega}}, \quad \forall (i, j) \in \Omega,
\end{equation}
where $(i,j)$ denotes pixel coordinates, $W(i,j) \in W$ represents the weight value at $(i,j)$, $\Delta D(i,j) = |D(i,j) - \hat{D}(i,j)|$ defines the depth discrepancy at $(i,j)$, and $\Omega$ is the image domain.
Using weight map $W$, an adaptive thresholding mechanism extracts high-weight trust regions $\mathcal{H} \subset \Omega$, prioritized due to their higher reliability, as shown in Fig.~\ref{fig:weightDisplay}. 
These weights simultaneously quantify feature importance to optimize constraint contributions in system optimization. 
The trust region selection criterion is:
\begin{equation}
    \mathcal{H} = \{ (i, j) \in \Omega \mid W(i, j) \geq \theta \},
\end{equation}
where $\theta$ is a set threshold, and the region above this threshold is the trust region.
\subsubsection{Texture Feature Decoupling}
\begin{figure}[b!]
    \centering
    \includegraphics[width=\columnwidth]{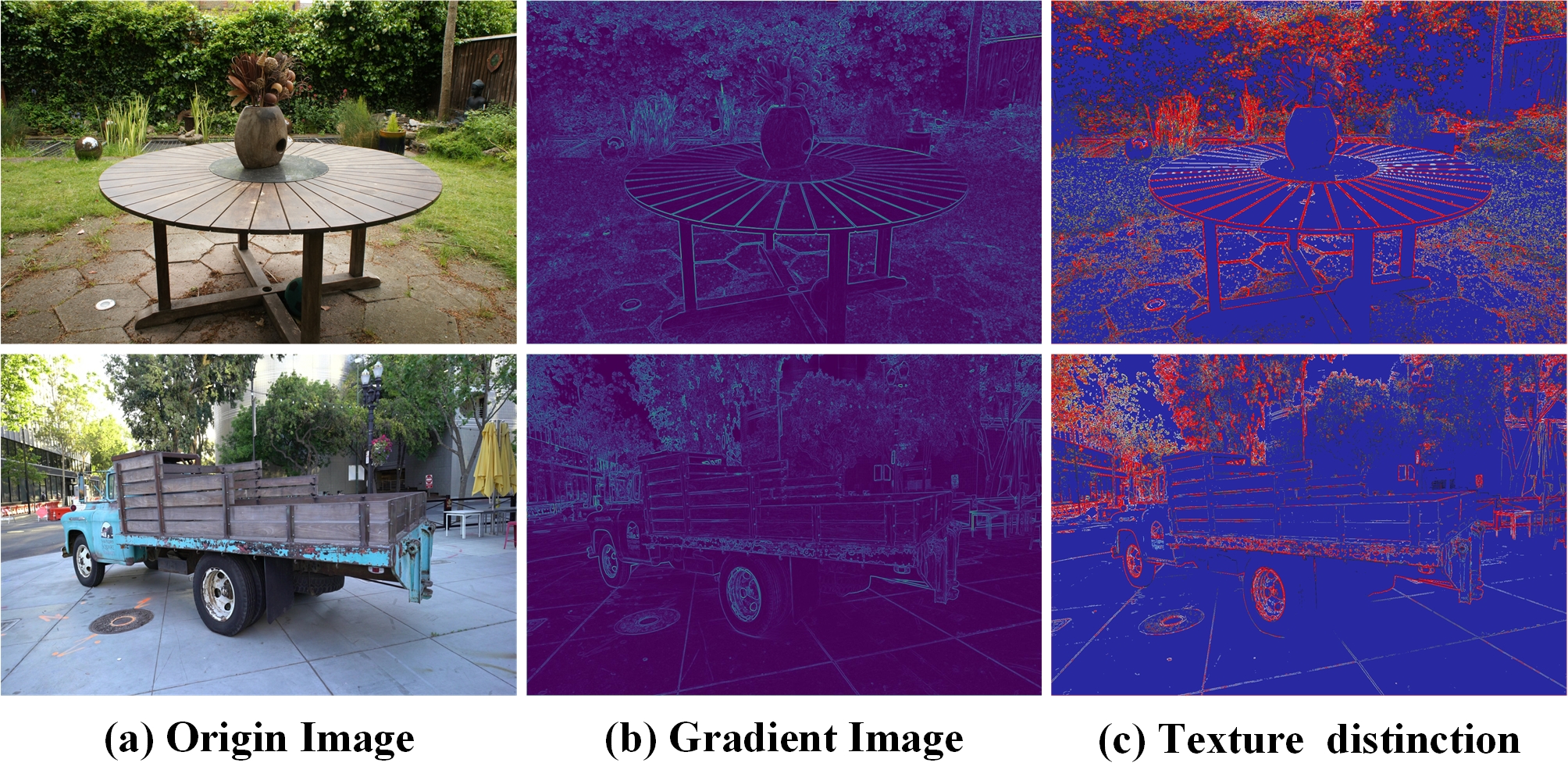}
    \caption{Calculate the gradient of the original image and classify the scene into texture-rich and texture-less regions according to the threshold. (a) Original RGB image. (b) Calculated gradient of the image, with the blue bias representing a flatter gradient region and the white bias a higher gradient region. (c) According to the threshold screening, the red part is a texture-rich region and the blue part is a texture-less region.}
    \label{fig:gradientDisplay}
\end{figure}

The image gradient is utilized to dynamically divide the texture-rich and texture-deficient regions of the scene. For the input image $I$, the horizontal gradient $G_x$ and vertical gradient $G_y$ are computed using the Sobel kernel~\cite{Sobel}:
\begin{equation}
    G_x(i,j) = \frac{\partial I(i,j)}{\partial x} \ast K_{x}, \quad G_y(i,j) = \frac{\partial I(i,j)}{\partial y} \ast K_{y},
\end{equation}
where $K_{x}$ and $K_{y}$ represent the Sobel operator in the horizontal and vertical directions, respectively. The final gradient magnitude can be calculated from the Euclidean parameter of $G_{x}(i,j)$ and $G_{y}(i,j)$:
\begin{equation}
    G(i, j) = \sqrt{G_x(i, j)^2 + G_y(i, j)^2}.
\end{equation}
Since the Sobel operator incorporates smoothing and computes gradients based on intensity differences, it effectively suppresses noise and is robust to global illumination variations.
To achieve robust dynamic segmentation, we set the threshold value $\tau$ to the 75th percentile of the gradient magnitude distribution. 
Consequently, pixels with gradients exceeding $\tau$ are classified as texture-rich regions $\mathcal{R}$, while those with lower gradients are assigned to texture-less regions $\mathcal{B}$, as shown in Fig. \ref{fig:gradientDisplay}. 
The detailed dynamic process is described as:
\begin{equation}
    p(i,j) \in 
\begin{cases} 
\mathcal{R} & \text{if } G(i,j) \geq \tau \\ 
\mathcal{B} & \text{otherwise}
\end{cases},
\end{equation}
where $p(i,j)$ represents the pixel.

\subsubsection{Single View Branch Normal Constraints}
In the ~\cite{chen2024pgsr} method, the original single-view constraints are constructed by weighting the rendered normal map by the power-of-five image gradient and the difference in unbiased depth.
The method has some limitations: in texture-less regions, the image gradient tends to 0 due to missing texture features, and the contribution of the normal constraint term decays.
Meanwhile, the traditional normal constraint term is sensitive to noise and prone to geometric ambiguity, leading to holes or floating artifacts on the object's surface.
To address the above problems, we propose a sub-area adaptive constraint method. 
Different normal constraints are designed in texture-rich and texture-less regions, and the weight $W$ is introduced in the texture-rich region to adaptively regulate the importance, and the robustness of the system is enhanced by the regional differentiation process.

In the reliable texture-rich region, based on the principle of orthogonality, the orthogonal constraint term $\mathcal{L}_{cross}$ between the depth gradient and the normal direction is introduced, which has the mathematical form:
\begin{equation}
\begin{split}
    \mathcal{L}_{cross} &= \frac{1}{|\mathbb{I}_{\mathcal{R} \cap \mathcal{H}}|} \sum_{i,j \in \mathbb{I}_{\mathcal{R} \cap \mathcal{H}}} \left|  \frac{\partial (\nabla D)}{\partial x} \cdot N_y - \frac{\partial (\nabla D)}{\partial y} \cdot N_x \right|.  \label{eq:L_cross_L_cross}
\end{split}
\end{equation}
Denote the system constraint under single-view as
\begin{equation}
\begin{split}
    \mathcal{L}_{svn} &= \frac{1}{|\mathbb{I}_{\mathcal{R} \cap \mathcal{H}}|} \sum_{(i,j) \in \Omega} \mathbb{I}_{\mathcal{R} \cap \mathcal{H}}(i,j) W(i,j) \left( \Delta N \right) + \lambda_1 \mathcal{L}_{cross},
    \label{eq:L_cross_L_svn}
\end{split}
\end{equation}where $\lambda_1$ weights the internal orthogonality constraint $\mathcal{L}_{cross}$;
\begin{figure}[b!]
    \centering
    \includegraphics[width=\columnwidth]{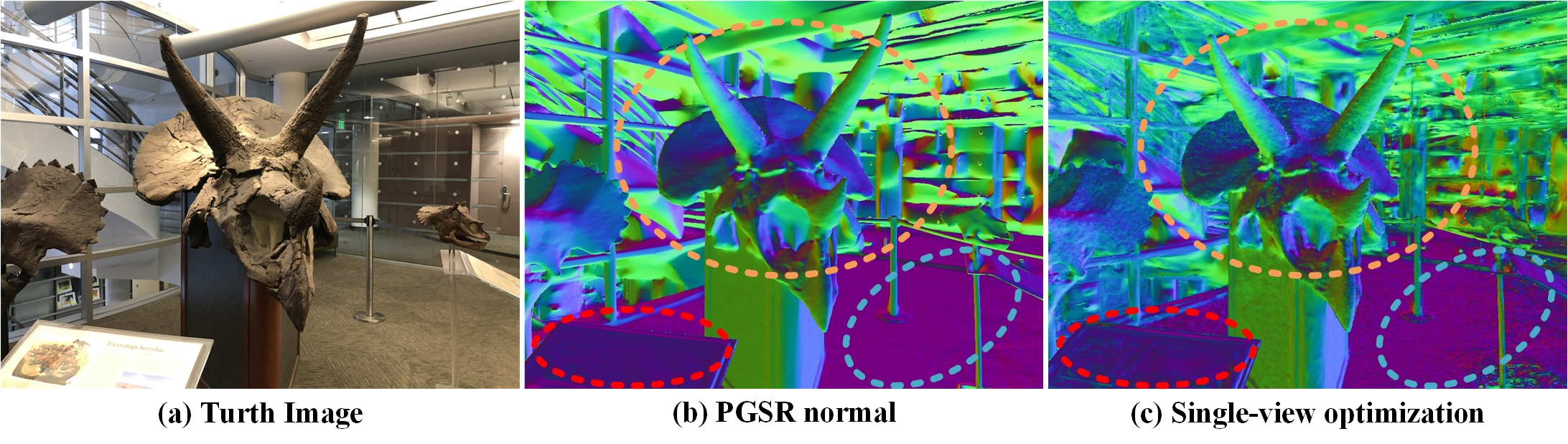}
    \caption{Analysis of reconstruction results based on normal maps. (a) Ground-truth RGB image; (b) Normal map reconstructed by the baseline PGSR algorithm; (c) Normal map reconstructed using the proposed single-view constrained optimization strategy. Comparative results demonstrate that our method yields more accurate geometric reconstruction than the baseline, particularly in preserving fine-grained details of real-world scenes.}
    \label{fig:single-view_normal}
\end{figure}
$\mathbb{I}_{\mathcal{R} \cap \mathcal{H}}(i,j) \subset \Omega$ designates the pixel set within texture-abundant trust regions, in which $|\mathbb{I}_{\mathcal{R} \cap \mathcal{H}}(i,j)|$ represents the regional pixel cardinality; $\Delta N = \| N_d - N \|_1$ denotes the difference between the normal map $N_d$ and the rendered normal obtained through unbiased depth, while $\nabla D$ denotes the gradient field computed from the discrepancy between unbiased depth and rendered depth values. The geometric surface undergoes optimization through trust-region normal constraints, with concurrent construction of gradient fields incorporating depth-normal orthogonality constraints to mitigate texture misinterpretation.
System constraint terms are refined via adaptive contribution weighting, ensuring high-frequency normal orientations maintain geometric fidelity while suppressing gradient conflict-induced over-regularisation artifacts.

In textured-less regions, a total variation (TV) regularization~\cite{liang2024gs} term weighted by color similarity is used:
\begin{equation}
  \begin{gathered}
    \Delta_{i,j}^{N} = \exp\left(-\left|I_{i,j} - I_{i-1,j}\right|\right) \left(N_{i,j} - N_{i-1,j}\right)^2 \\
    \quad + \exp\left(-\left|I_{i,j} - I_{i,j-1}\right|\right) \left(N_{i,j} - N_{i,j-1}\right)^2, \\
    TV_{normal} = \frac{1}{|\mathcal{B}|} \sum_{(i,j) \in \Omega} \alpha(i,j) \Delta_{i,j}^{N}.
  \end{gathered}
\end{equation}Let $|\mathcal{B}|$ represent the pixel cardinality of set $\mathcal{B}$, where $\alpha(i,j) \in \mathcal{B}$ indexes pixel positions. $I_{i,j}$ and $N_{i,j}$ denote the RGB intensity and surface normal vector at coordinate $(i,j)$, respectively. 
A color similarity-weighted exponential decay operator adaptively modulates the smoothing intensity.
Within texture-impoverished regions, color consistency-driven geometric smoothness constraints simultaneously mitigate texture scarcity-induced reconstruction errors and suppress stochastic noise propagation. 
The chromatic similarity weighting preserves physically plausible discontinuities while preventing edge over-smoothing, thereby enhancing both geometric completeness and reconstruction stability in texture-less domains through discontinuity-aware regularization.

\begin{algorithm}
\caption{Texture-aware Normal Loss}\label{singleAlgorithm}
\KwIn{
    \begin{tabular}[t]{r@{ : }l}
        $D,\hat{D},M_d,N,I_{gt}$& Geometric information \\
        $\lambda_1,\lambda_2, \tau,\theta$& Weight and threshold \\
    \end{tabular}
}

\KwOut{
    \begin{tabular}[t]{r@{ : }l}
        $\mathcal{L}_{svgeo}$& Texture-aware normal loss 
    \end{tabular}
}

\For{\(t = 0\) \KwTo \(k\)}
{
    $W \gets  \text{DeltaDepthWeight}(D, \hat{D})$\
    $G \gets  \text{ComputeImageGradient}(I_{gt})$\ \\
    \For{each $(i,j) \in \mathbb{I}_{\mathcal{R} \cap \mathcal{H}}$}
    {
        $\mathcal{H} \gets  \{(i,j) \mid W(i,j) \geq \theta\}$\
        $\mathcal{R} \gets  \{(i,j) \mid G(i,j) \geq \tau\}$, \quad $\mathcal{B} \gets  \Omega \setminus \mathcal{R}$\
        $\nabla D(i,j) \gets  \text{DeltaDepthGrad}(D(i,j), \hat{D}(i,j))$\
        $\mathcal{L}_{tex} \gets  \text{SVRTexNorLoss}(N_d(i,j),N(i,j))$\
        $\mathcal{L}_{cross} \gets  \text{DeltaDepthNorCross}(\nabla D, N)$\
        $\mathcal{L}_{svn} += \mathcal{L}_{tex} + \lambda_1 \mathcal{L}_{cross}$\
    }
    
    \For{each $(i,j) \in \mathcal{B} $}
    {
        $TV_{normal}(i,j) += \text{SVWTextNorLoss}(I_{gt},N)$\
    }
    $\mathcal{L}_{svn} \gets  \frac{1}{|\mathcal{R} \cap \mathcal{H}|} \mathcal{L}_{svn} , \quad TV_{normal} \gets \frac{1}{|\mathcal{B}|}  TV_{normal}$\ \\
    $\mathcal{L}_{svgeo} \gets \mathcal{L}_{svn} + \lambda_2 TV_{normal}$\
}
\Return $\mathcal{L}_{svgeo}$\
\end{algorithm}

This framework implements a regionally-specialized constraint paradigm, enforcing strict constraints via dual-criteria evaluation of trust regions and texture abundance for primary texture-rich domains while employing streamlined trust region assessment for spatially limited texture-impoverished areas. 
The resultant single-view geometric constraint formulation is expressed as:
\begin{equation}
    \mathcal{L}_{svgeo}=\mathcal{L}_{svn}+\lambda_2 TV_{normal},
\end{equation}
where $\lambda_2$ regulates the intensity of the total variation term $TV_{normal}$.
To validate the effectiveness of the dual-branch constraint strategy based on regional texture feature differences, a comparative analysis of normal maps rendered by the optimized method and the baseline PGSR algorithm is presented in Fig.~\ref{fig:single-view_normal}. The results demonstrate that the proposed method significantly enhances geometric detail reconstruction in both texture-rich regions (dinosaur head) and texture-less regions (ground and introductory sign), thereby improving geometric fidelity in real-world scene representation.

\subsection{Geometry-guided Multi-view Consistency Constraints}
\label{app:geometry_guided}
\begin{figure}[t!]
    \centering
    \includegraphics[width=\columnwidth]{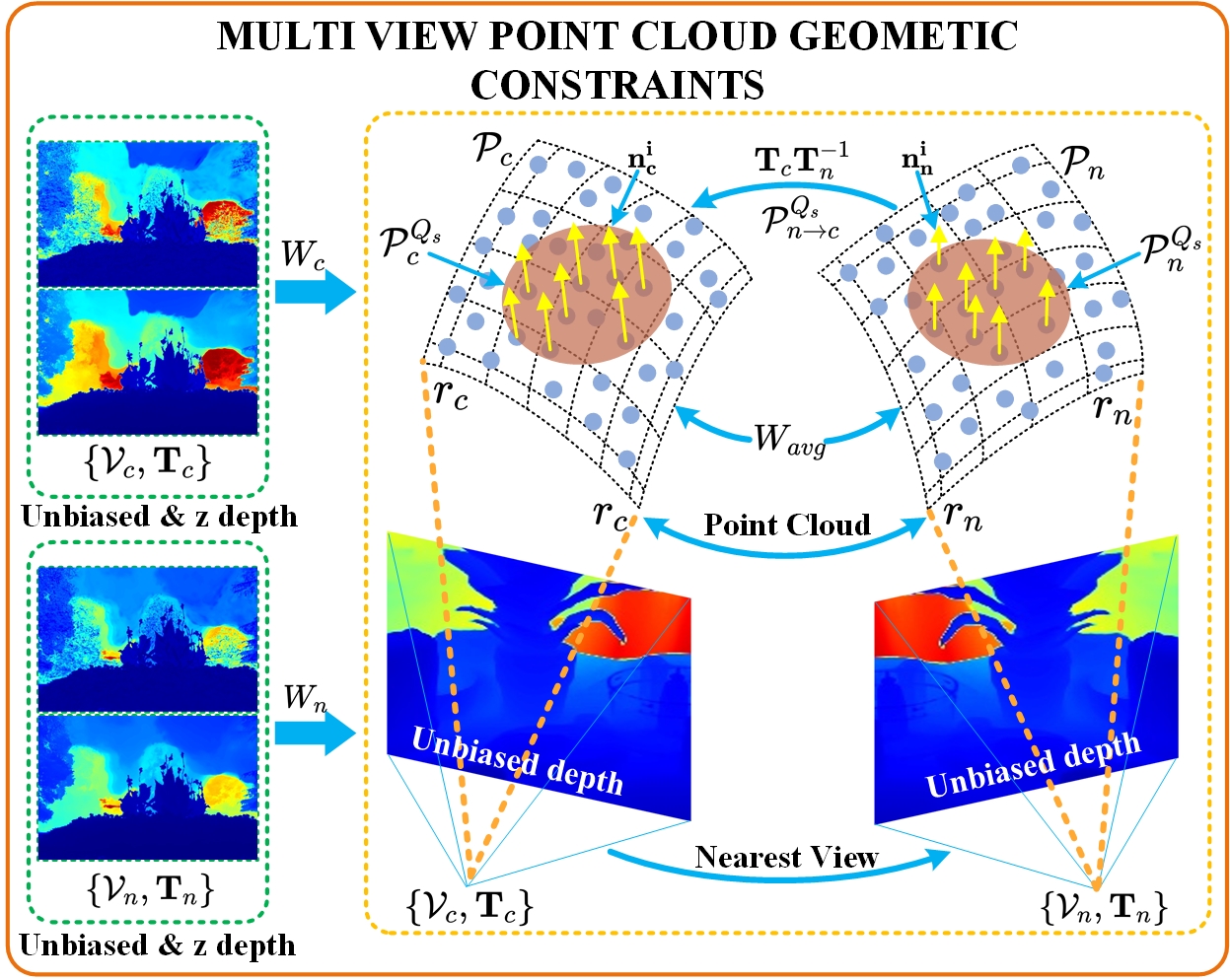}
    \caption{Guaranteeing multi-view consistency to minimize spatial artifacts through cross-view spatial point cloud geometric constraints. Where $r_c$ and $r_n$ are camera rays, $\mathcal{P}_c,\mathcal{P}_n$ are depth-mapped point cloud points, $\mathcal{P}^{Q_s}_c,\mathcal{P}^{Q_s}_n$ are point clouds sampled based on the global weights $W_{avg}$,$\mathbf{n_c^i},\mathbf{n}_n^i$ are the surface normal vectors of the point cloud in two frames. The current view $\{\mathcal{V}_c,\mathbf{T}_c\}$ and the neighbouring view $\{\mathcal{V}_n,\mathbf{T}_n\}$ are obtained first, and the two frames of the point cloud are generated from the depth map, and then the block of the point cloud is sampled based on the depth difference weights and the global weights to complete the constraints on the surface normal of the three-dimensional space.}
    \label{fig:Multi-view}
\end{figure}

Under multi-view constraints, the 3D object surface constraints are constructed based on the principle of re-projection error minimization, and the algorithm framework is shown in Fig.~\ref{fig:Multi-view}. 
The specific workflow proceeds as follows: During the system's rendering and reconstruction phases, the nearest neighboring view to the current perspective is first identified. 
Depth information from both views is then utilized to project corresponding pixels into 3D space, generating complementary point clouds. 
Leveraging the relative pose transformation between the current and neighboring views, the neighboring point cloud is transformed into the current view’s coordinate system.
Under ideal reconstruction and rendering conditions, these aligned point clouds would exhibit perfect spatial congruence in geometric space, accompanied by identical surface normal orientations. 
Global confidence weights are derived by integrating view-specific weighting metrics from both perspectives. 
High-confidence point cloud samples are subsequently selected to establish cross-view geometric consistency constraints, thereby enhancing the framework’s reconstruction fidelity and rendering precision.

\subsubsection{Weight-guided Geometric Sampling Strategy}
In this method, 3D point cloud data $\mathcal{P}_c, \mathcal{P}_{n}$ are based on the current view $\mathcal{V}_c$ and its spatially nearest neighboring view $\mathcal{V}_n$. 
The distance depth maps $D_1, D_2$ and unbiased depth maps $\hat{D}_1,\hat{D}_2$ are output by the rendering operation. In order to reduce the computational complexity, a geometric sampling strategy based on global weight guidance is proposed: firstly, the global weight $W_{avg}$ is obtained by fusing the cross-view weights $W_c, W_n$, which are calculated as follows:
\begin{equation}
    W_{avg}(i,j) = \beta W_c(i,j) + (1 - \beta) W_n(i,j).
\end{equation}Let $W_c(i,j) \in \mathbb{R}^{H \times W}$ and $W_n(i,j) \in \mathbb{R}^{H \times W}$ denote the view-specific confidence weights computed for the current view and its nearest neighbouring view, respectively, with $\beta=0.5$ balancing their contributions.
The global confidence weights $W_{avg}(i,j)$, derived as a convex combination of $W_c$ and $W_n$, quantify the joint reliability of cross-view geometric reconstruction. 
To harmonize computational efficiency with reconstruction fidelity, a confidence-driven adaptive sampling strategy is proposed to selectively retain stable geometric primitives from the 3D point cloud. 
This strategy initiates by constructing a binary geometric validity mask $\mathcal{M}_d \in \{0,1\}^{H \times W}$ through projection consistency verification, defined as:
\begin{equation}
    \mathcal{M}_d(i,j) = 
    \begin{cases} 
        1, & (u_i,v_i) \in [0,W)\times[0,H) \ \land \ z_i \geq \epsilon_d \\ 
        0, & \text{otherwise}
    \end{cases} ,
\end{equation}
where $(u_i,v_i)$ is the normalised projected coordinates of the 3D point cloud in the image plane, $ H$ and $ W$ are the image size, and $\epsilon_d$ is the depth threshold, set to 0.1 m following the default setting of the original framework~\cite{chen2024pgsr}, to filter near-field noise from the camera. 
The set of candidate regions is constructed by logically operating the global weights $W_{avg}$ with the mask $\mathcal{M}_d$:
\begin{equation}
    \mathcal{Q} = \{ (i,j) \in \Omega \mid \mathcal{M}_d(i,j) = 1 \land W_{avg}(i,j) \geq \gamma \} ,
\end{equation}
where $\gamma$ is the adaptive dynamic threshold, set to $30\%$ of $W_{avg}$ to dynamically filter low-confidence regions and ensure the quality of the candidate set $\mathcal{Q}$. Subsequently, the top sample rate $S$ of high-weighted points are sampled in descending order of confidence:
\begin{equation}
    \label{eq:16}
    \mathcal{Q}_s=\{(i_s,j_s) \in \mathcal{Q} | \text{rank}(W_{avg}(i_s,j_s)) \leq S\} .
\end{equation}
The operator $\text{rank}(\cdot)$ in Eq.~\ref{eq:16}  sorts the weights in descending order. 
The global-weight-driven geometric sampling strategy employs prioritized probabilistic sampling within high-confidence regions, thereby enforcing geometric constraints exclusively on stable geometric primitives.
Crucially, this strategy functions as an implicit occlusion-aware mechanism by automatically excluding regions where significant depth discrepancies yield low weights $W$, thereby preventing erroneous geometric associations from degrading the optimization.

\subsubsection{Cross-view Sampled Point Cloud Geometry Constraints}
Following the weight-driven geometric sampling strategy, high-confidence point sets $\mathcal{P}_c$ and their nearest-neighbor counterparts $\mathcal{P}_n$ are extracted from the current-view point cloud $\mathcal{P}_{c}^{\mathcal{Q}s}$ and the neighboring-view point cloud $\mathcal{P}_{n}^{\mathcal{Q}_s}$. 
To establish geometrically consistent constraints using the sampled 3D point clouds, both sets must reside within a unified coordinate system. Consequently, $\mathcal{P}_{n}^{\mathcal{Q}_s}$ undergoes a rigid transformation to align with the current-view coordinate frame.
Let $\mathbf{T}_c, \mathbf{T}_n \in \mathrm{SE}(3)$ denote the poses of the current and neighboring views, respectively, within the world coordinate system:
\begin{equation}
    \mathbf{T}_c = \begin{bmatrix}
    \mathbf{R}_c & \mathbf{t}_c \\
    \mathbf{0}^T & 1
    \end{bmatrix},
    \ \ \
    \mathbf{T}_n = \begin{bmatrix}
    \mathbf{R}_n & \mathbf{t}_n \\
    \mathbf{0}^T & 1
    \end{bmatrix} ,
\end{equation} 
where $\mathbf{R}_c,\mathbf{R}_n \in SO(3)$ is the rotation matrix and $\mathbf{t}_c,\mathbf{t}_n \in \mathbb{R}^3$ is the translation vector. 
According to the relative positional transformation $\mathbf{T_c}\mathbf{T_n}^{-1}$, the nearest-neighbor frame point $\mathbf{p}_{n} $ is transformed to the current frame coordinate system, and the transformed point cloud is:
\begin{equation}
    \mathcal{P}_{n\to c}^{\mathcal{Q}_s} = \Biggl\{ 
    \mathbf{\Pi}_3\Bigl( \mathbf{T}_c \mathbf{T}_n^{-1} \tilde{\mathbf{p}}_n \Bigr) 
    \,\Bigg\vert\, 
    \forall \tilde{\mathbf{p}}_n = [\mathbf{p}_n^\top \; 1]^\top,\; \mathbf{p}_n \in \mathcal{P}_n^{\mathcal{Q}_s} 
\Biggr\} ,
\end{equation}where $\mathbf{\Pi}_3(\cdot)$ projects homogeneous coordinates to three-dimensional coordinates, and $\tilde{\mathbf{p}} \in \mathbb{R}^4$ is the homogeneous coordinate representation of the point $\mathbf{p}_{n}$.
\begin{table*}[b]
    \centering
    \caption{Quantitative analysis of the 3D scene reconstruction accuracy of the algorithm using chamfer distance (mm)$\downarrow$ on the DTU dataset. "Red", "Orange", and "Yellow" indicate the best, second-best, and third-best results, respectively.}
    \label{tab:quantitativeDTU}
    \resizebox{\textwidth}{!}{
    \begin{tabular}{cccccccccccccccccc}
        \toprule
        \textbf{Methods} & \textbf{24} & \textbf{37} & \textbf{40} & \textbf{55} & \textbf{63} & \textbf{65} & \textbf{69} & \textbf{83} & \textbf{97} & \textbf{105} & \textbf{106} & \textbf{110} & \textbf{114} & \textbf{118} & \textbf{122} & \textbf{Mean} & \textbf{Mean Time}\\
        \midrule
        NeRF~\cite{mildenhall2021nerf} & 1.90 & 1.60 & 1.85 & 0.58 & 2.28 & 1.27 & 1.47 & 1.67 & 2.05 & 1.07 & 0.88 & 2.53 & 1.06 & 1.15 & 0.96 & 1.49 & $>$ 0.8h\\
        VolSDF~\cite{yariv2021volume} & 1.14 & 1.26 & 0.81 & 0.49 & 1.25 & 0.70 & 0.72 & 1.29 & \cellcolor{yellow!50}1.18 & 0.70 & 0.66 & 1.08 & \cellcolor{yellow!50}0.42 & 0.61 & 0.55 & 0.86 & $>$ 0.8h\\
        NeuS~\cite{wang2021neus} & 1.00 & 1.37 & 0.93 & 0.43 & 1.10 & \cellcolor{yellow!50}0.65 & \cellcolor{orange!50}0.57 & 1.48 & \cellcolor{orange!50}1.09 & 0.83 & \cellcolor{yellow!50}0.52 & 1.20 & \cellcolor{orange!50}0.35 & \cellcolor{yellow!50}0.49 & 0.54 & 0.84 & $>$ 8.5h\\
        \midrule
        2DGS~\cite{huang20242d} & 0.52 & \cellcolor{yellow!50}0.82 & \cellcolor{orange!50}0.35 & 0.42 & 0.93 & 0.97 & 0.82 & 1.23 & 1.24 & \cellcolor{yellow!50}0.64 & 0.68 & 1.27 & \cellcolor{yellow!50}0.42 & 0.67 & 0.48 & 0.76 & \cellcolor{orange!50}0.15h \\
        GOF~\cite{yu2024gaussian} & 0.53 & 0.83 & 0.40 & 0.38 & 1.35 & 0.82 & 0.79 & 1.26 & 1.30 & 0.69 & 0.71 & 1.37 & 0.52 & 0.64 & 0.51 & 0.81 & 0.58h \\
        RaDe-GS~\cite{zhang2024rade} & \cellcolor{yellow!50}0.43 & \cellcolor{orange!50}0.75 & \cellcolor{red!50}\textbf{0.34} & \cellcolor{yellow!50}0.37 & \cellcolor{yellow!50}0.84 & 0.72 & \cellcolor{yellow!50}0.67 & \cellcolor{yellow!50}1.20 & 1.24 & \cellcolor{yellow!50}0.64 & 0.62 & \cellcolor{yellow!50}0.85 & \cellcolor{orange!50}0.35 & 0.66 & \cellcolor{yellow!50}0.47 & \cellcolor{yellow!50}0.68 & \cellcolor{red!50}\textbf{0.13h}\\
        PGSR~\cite{chen2024pgsr} & \cellcolor{orange!50}0.37 & \cellcolor{red!50}\textbf{0.55} & 0.42 & \cellcolor{orange!50}0.35 & \cellcolor{orange!50}0.78 & \cellcolor{orange!50}0.58 & \cellcolor{red!50}\textbf{0.49} & \cellcolor{orange!50}1.08 & \cellcolor{red!50}\textbf{0.64} & \cellcolor{orange!50}0.59 & \cellcolor{orange!50}0.48 & \cellcolor{orange!50}0.53 & \cellcolor{red!50}\textbf{0.30} & \cellcolor{orange!50}0.37 & \cellcolor{orange!50}0.35 & \cellcolor{orange!50}0.53 & \cellcolor{yellow!50}0.28h \\
        Ours & \cellcolor{red!50}\textbf{0.34} & \cellcolor{red!50}\textbf{0.55} & \cellcolor{yellow!50}0.37 & \cellcolor{red!50}\textbf{0.34} & \cellcolor{red!50}\textbf{0.77} & \cellcolor{red!50}\textbf{0.55} & \cellcolor{red!50}\textbf{0.49} & \cellcolor{red!50}\textbf{1.04} & \cellcolor{red!50}\textbf{0.64} & \cellcolor{red!50}\textbf{0.58} & \cellcolor{red!50}\textbf{0.47} & \cellcolor{red!50}\textbf{0.48} & \cellcolor{red!50}\textbf{0.30} & \cellcolor{red!50}\textbf{0.36} & \cellcolor{red!50}\textbf{0.33} & \cellcolor{red!50}\textbf{0.51} & 0.45h \\
        \bottomrule
    \end{tabular}
    }
\end{table*}
The $\mathcal{P}_{c}^{\mathcal{Q}_s}$ and $\mathcal{P}_{n \rightarrow c}^{\mathcal{Q}_s}$ are constrained to the surface normals of the two frames of the sampled point cloud under the same coordinate system based on the assumption of the local plane. 
Domain facets of $3 \times 3$ are extracted at corresponding points in the sampled point clouds $\mathcal{P}_{c}^{\mathcal{Q}_s}$ and $\mathcal{P}_{n}^{\mathcal{Q}_s}$ and are denoted as $\{\mathbf{x}_1, \ldots, \mathbf{x }_9\}$ and $\{\mathbf{y}_1, \ldots, \mathbf{y}_9\}$, respectively. 
Principal Component Analysis (PCA)~\cite{PCA_algorithm} is utilised to solve for the normal vector of each point in $ \mathcal{P}_{n}^{\mathcal{Q}_s}$ and $ \mathcal{P}_{n \rightarrow c}^{Q_s}$, denoted as $ \mathbf{n}_c^{\mathcal{Q}_s} $ and $ \mathbf{n}_n^{\mathcal{Q}_s} $. The normal vector of each point cloud is calculated as:
\begin{equation}
\begin{split}
    \mathbf{n}_c^i = \arg \min _{\substack{\mathbf{v} \in \mathbb{R}^3 \\ \|\mathbf{v}\|=1}} \mathbf{v}^\top \sum_{k=1}^M (\mathbf{x}_k - \mu_c)(\mathbf{x}_k - \mu_c)^\top \mathbf{v} , \\
    \mathbf{n}_n^i = \arg \min _{\substack{\mathbf{v} \in \mathbb{R}^3 \\ \|\mathbf{v}\|=1}} \mathbf{v}^\top \sum_{k=1}^M (\mathbf{y}_k - \mu_n)(\mathbf{y}_k - \mu_n)^\top \mathbf{v} ,
\end{split}
\end{equation}where $\mathbf{n}_c^i \in \mathbf{n}_c^{\mathcal{Q}_s},\mathbf{n}_n^i \in \mathbf{n}_n^{\mathcal{Q}_s} $ denote the two-frame point cloud surface normal vectors; $\mu_c = \frac{1}{M} \sum_ {k=1}^M \mathbf{x}_k $ and $\mu_n = \frac{1}{M} \sum_{k=1}^M \mathbf{y}_k$ denote the mean value of localized surface sheets of the current view and the transformed view, respectively ($M=9$); $\mathbf{v}$ is the direction of the unit normal vector constraints of the PCA solution ($||\mathbf{v}||=1$). 
To quantify the local geometric stability, the normalized curvature is defined based on the PCA eigenvalues. The higher the curvature, the less flat the plane is, and the more unstable the geometric estimate is, so an exponential decay weight function is constructed:
\begin{equation}
    w_\kappa^i = \exp \left( -10 \cdot \frac{\eta_c^3}{\sum_{k=1}^3 \eta_c^k} \right), \quad \eta_c^1 \geq \eta_c^2 \geq \eta_c^3 ,
\end{equation}where $\eta_c^k$ represents the eigenvalue of the local surface sheet in the current view, and $w_\kappa^i \in w_\kappa$ denotes the local surface sheet curvature.
Using the curvature to weight the surface normal constraints, the weighted cosine similarity loss term is:
\begin{equation}
    \mathcal{L}_{mvgeo} = \frac{1}{|\mathcal{K}|}\sum_{i \in |\mathcal{K}|} w_\kappa^i  \cdot \left( 1 - \left| {\mathbf{n}_c^{i}}^T \mathbf{n}_n^{i} \right|\right) ,
\end{equation}
where $\mathcal{K}$ denotes the set of point cloud surface facets, $|\mathcal{K}|$ denotes the number of surfaces, and $w_\kappa^i \in w_\kappa$ denotes the curvature. The process of constructing multi-view consistency constraints based on geometric guidance is shown in Algorithm~\ref{multiAlgorithm}.
\begin{algorithm}
\caption{Multi-view Geometry Constraints}\label{alg:multi_view_spatial_loss}\label{multiAlgorithm}
\KwIn{
    \begin{tabular}[t]{r@{ : }l}
        $\mathcal{V}_c, \mathcal{V}_n$& Camera perspective \\
        $\mathbf{T}_c, \mathbf{T}_n$& Camera pose \\ 
        $\mathcal{G}$& Gaussian Representation \\
    \end{tabular}
}

\KwOut{
    \begin{tabular}[t]{r@{ : }l}
        $\mathcal{L}_{mvgeo}$& Multi-view Geometry loss
    \end{tabular}
}

{
    $\{D_c, \hat{D}_c, N_c\} \gets \text{Render}(\mathcal{V}_c, \mathcal{G})$\
    $\{D_n, \hat{D}_n, N_n\} \gets \text{Render}(\mathcal{V}_n, \mathcal{G})$\
    $\{\mathcal{P}_c, \mathcal{P}_{n}\} \gets \text{GetPointsFromDepth}(\mathcal{V}_c, \mathcal{V}_n, \hat{D}_c, \hat{D}_n) $\
    $\{W_c, W_n \} \gets \text{DeltaDepthWeight}(D_c, \hat{D}_c, D_n, \hat{D}_n)$\
    $W_{avg} \gets \beta W_c + (1 - \beta) W_n$ \
    $\mathcal{Q} \gets \{ (i,j) \in \Omega \mid \mathcal{M}_d(i,j) \gets 1 \land W_{avg}(i,j) \geq \gamma \}$\
    $\mathcal{Q}_s \gets \{(i_s,j_s) \in \mathcal{Q} \mid \text{rank}(W_{avg}(i_s,j_s)) \leq S \}$\
    $\mathcal{P}_{n\to c}^{\mathcal{Q}_s} \gets \Biggl\{\mathbf{\Pi}_3\Bigl( \mathbf{T}_c \mathbf{T}_n^{-1} [\mathbf{p}_n^\top \; 1]^\top \Bigr), \mathbf{p}_n \in \mathcal{P}_n^{\mathcal{Q}_s} \Biggl\}$\
 
}


\For{each $p_c \in \mathcal{P}_{c}^{\mathcal{Q}_s}$ and $p_n \in \mathcal{P}_{n}^{\mathcal{Q}_s}$}
{
    \tcp{Compute eigenvalues and eigenvectors of local patch} 
    $(\lambda_c^1,\lambda_c^2,\lambda_c^3; v_c^1,v_c^2,v_c^3) \gets \text{PatchPCA}(p_c, P=3)$ \\
    $(\lambda_n^1,\lambda_n^2,\lambda_n^3; v_n^1,v_n^2,v_n^3) \gets \text{PatchPCA}(p_n, P=3)$ \\
    \tcp{Select eigenvector corresponding to smallest eigenvalue}
    $\mathbf{n}_c^i \gets v_c^1 \quad (\lambda_c^1 \le \lambda_c^2 \le \lambda_c^3)$ \\
    $\mathbf{n}_n^i \gets v_n^1 \quad (\lambda_n^1 \le \lambda_n^2 \le \lambda_n^3)$  \\
    $\kappa^i \gets \dfrac{\lambda_c^3}{\sum_{k=1}^3 \lambda_c^k}, 
    \quad w_\kappa^i \gets \exp(-10 \kappa^i)$   \\
    $\mathcal{L}_{mvgeo} += w_\kappa^i \cdot 
    \left(1 - \left| {\mathbf{n}_c^i}^\top \mathbf{n}_n^i \right| \right)$  \\
}

\Return $\mathcal{L}_{mvgeo}$\
\end{algorithm}
\begin{figure*}[t]
    \centering
    \includegraphics[width=\textwidth]{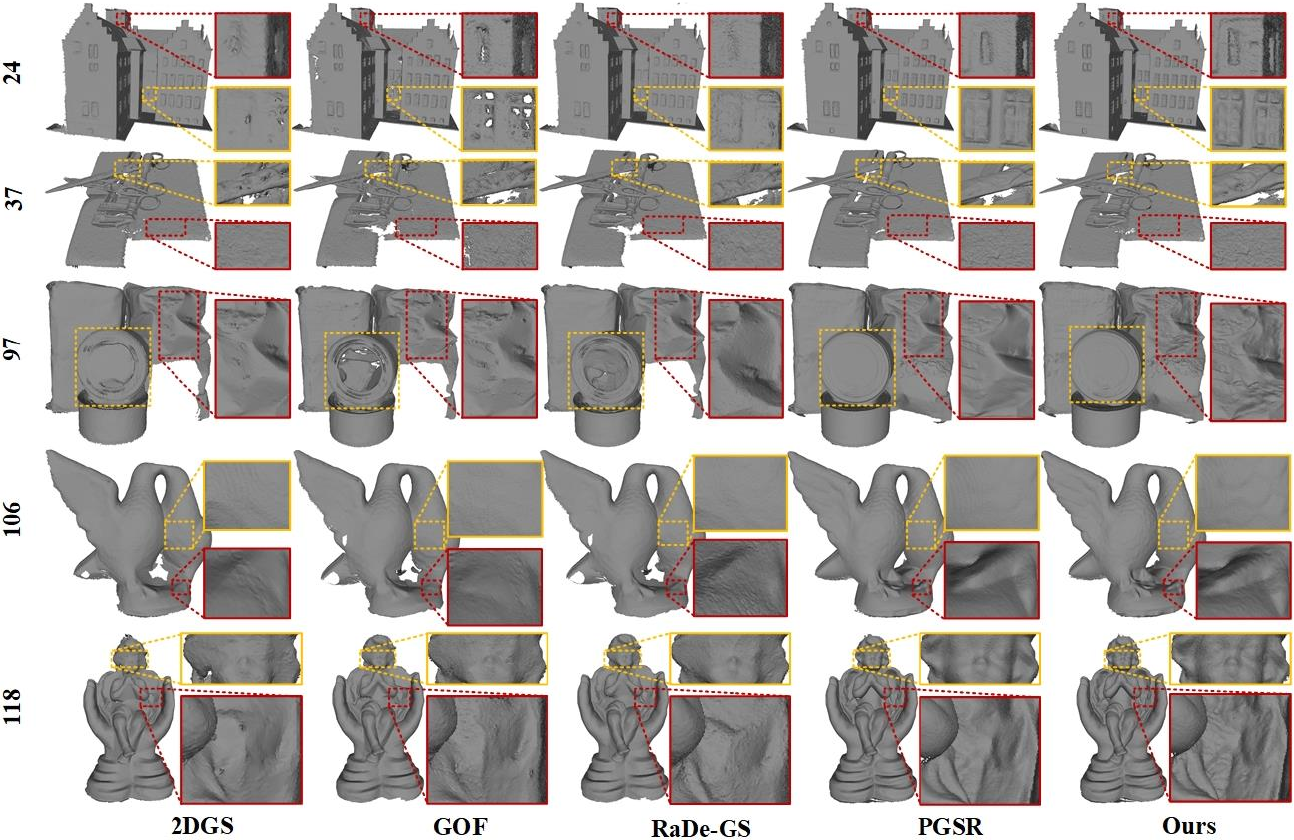} 
    \caption{Qualitative analysis of surface reconstruction on DTU. We compare with existing methods to better capture object surface details and more clearly represent object surface features.}
    \label{fig:reconstruction}
\end{figure*}
\subsection{Loss Functions}
\label{app:loss_functions}
To summarise, incorporating the RGB reconstruction loss $\mathcal{L}_{rgb}$, employed in 3D Gaussian Splatting, yields the final total loss term for the training process:
\begin{equation}
    \mathcal{L} = \mathcal{L}_{rgb} + \mathcal{L}_{svgeo} +  \lambda_3 \mathcal{L}_{mvgeo},
\end{equation}
where $\lambda_3>0$ denotes a weighting factor.
For the weight settings of the system constraints, for the photometric constraints, we keep the same weights as in 3DGS, with the innovative single-view geometric constraints weights $\lambda_1=0.05$ and $\lambda_2=0.01$, and the multi-view cross-view point cloud geometric constraints weights are set to $\lambda_3=0.001$.

\section{Experiements}
\label{app:experiements}
\subsection{Experimental Settings}
\textbf{Datasets and Evaluation Metrics:} 
To validate the effectiveness of the proposed method, experiments were conducted on three widely used datasets: DTU~\cite{jensen2014large}, Mip-NeRF360~\cite{barron2022mip}, and Tanks and Temples~\cite{knapitsch2017tanks}. 
The DTU dataset comprises 15 scenes that cover both reflective and shadowed areas; this dataset is commonly used to verify 3D reconstruction performance. 
The Mip-NeRF360 dataset provides 360-degree omnidirectional data across 9 indoor and outdoor scenes. 
Its long viewpoint trajectories, multi-scale geometric structures, and complex lighting variations rigorously test algorithms for continuous viewpoint synthesis and detail rendering. 
The Tanks and Temples dataset features large-scale real-world scenes with dynamic illumination and intricate occlusions, enabling a comprehensive evaluation of algorithmic robustness in unconstrained environments.
To quantify algorithm performance, Chamfer distance (CD) was used to measure 3D reconstruction quality; Peak Signal-to-Noise Ratio (PSNR), Structural Similarity Index (SSIM), and Learned Perceptual Image Patch Similarity (LPIPS) metrics were used to evaluate novel view synthesis performance.

\textbf{Baselines:}
To evaluate the superiority of this method, a comprehensive comparative analysis of 3DGS~\cite{kerbl20233d}, 2DGS~\cite{huang20242d}, GOF~\cite{yu2024gaussian}, RaDe-GS~\cite{zhang2024rade}, PGSR~\cite{chen2024pgsr}, and the proposed approach is performed on the Mip-NeRF360, Tanks and Temple, and  DTU datasets in terms of both novel view synthesis and 3D reconstruction accuracy.
Experimental comparisons are performed in various scenarios, integrating qualitative and quantitative analyses to provide comprehensive empirical support for the superiority of the method presented in this document.

\textbf{Implementation Details:} 
The framework is architected using PyTorch, C++, and CUDA technology stacks, with all experimental validations conducted on an NVIDIA GeForce RTX 4090 GPU featuring 24GB of VRAM. 
The system initialization phase employs COLMAP-generated input images, point clouds, and camera poses as foundational data. 
We implement a two-phase optimization strategy across 30,000 training iterations: initial 7,000 iterations prioritize parameter optimization through photometric constraints, while subsequent 23,000 iterations enhance model fidelity by incorporating single/multiview geometric constraints. 
Our hybrid densification module synergistically integrates the AbsGS~\cite{ye2024absgs} and GOF~\cite{yu2024gaussian} methodologies, with activation limited to the first 15,000 iterations. The remaining hyperparameters maintain consistency with the original 3D Gaussian Splatting implementation.

In the design of the geometric constraint mechanism, we set multiple thresholds to better regulate the system effect: the single-view scene is configured with a confidence threshold $\theta=0.8$ to screen valid regions; a texture feature difference threshold $\tau$, set to the 75th percentile of the image gradient histogram, for feature delineation; the multiview environment adopts a depth threshold $\epsilon_d=0.1$ to eliminate near-field noise and further optimizes the geometric constraints using $S=16$ with a random point cloud block sampling strategy.
The sensitivity analysis results of the above parameters are presented in Fig.~\ref{fig:appendix_theta_threshold} and \ref{fig:ablation_tau_threshold} of Appendix~\ref{app:Appendix_single_view_parameter}, as well as Fig.~\ref{fig:appendix_s_threshold} and Table~\ref{tab:appendix_s_table} in Appendix~\ref{app:Appendix_multi_view_parameter}.

\begin{table}[t]
    \centering
    \caption{Quantitative comparison of the Tanks and Temple dataset, comparing the F1-Score, the experimental results show that our method achieves optimal results in most scenarios.}
    \label{tab:quantativeTanks&Temple}
    \resizebox{\columnwidth}{!}{
    \begin{tabular}{cccccc}
        \toprule
         & 2DGS~\cite{huang20242d} & GOF~\cite{yu2024gaussian} & RaDe-GS~\cite{zhang2024rade} & PGSR~\cite{chen2024pgsr} & Ours \\
        \midrule
        Caterpillar   & 0.22 & \cellcolor{yellow!50}0.40 & 0.36 & \cellcolor{orange!50}0.41 & \cellcolor{red!50}\textbf{0.43} \\
        Courthouse    & 0.14 & \cellcolor{red!50}\textbf{0.28} & \cellcolor{orange!50}0.27 & 0.20 & \cellcolor{yellow!50}0.21 \\
        Ignatius      & 0.51 & 0.65 & \cellcolor{yellow!50}0.72 & \cellcolor{orange!50}0.77 & \cellcolor{red!50}\textbf{0.79} \\
        Meetingroom   & \cellcolor{yellow!50}0.05 & 0.06 & 0.04 & \cellcolor{red!50}\textbf{0.13} & \cellcolor{red!50}\textbf{0.13} \\
        Truck         & \cellcolor{yellow!50}0.16 & \cellcolor{yellow!50}0.16 & 0.14 & \cellcolor{orange!50}0.21 & \cellcolor{red!50}\textbf{0.22} \\
        \midrule
        Mean          & 0.22 & \cellcolor{yellow!50}0.31 & \cellcolor{yellow!50}0.31 & \cellcolor{orange!50}0.34 & \cellcolor{red!50}\textbf{0.36} \\
        \bottomrule
    \end{tabular}
    }
\end{table}

\begin{table*}[t]
    \centering
    \caption{Quantitative results of rendering quality for new view synthesis on the Mip-NeRF360 dataset. “Red”, “orange”, and “yellow” represent the best, second-best, and third-best results.}
    \label{tab:renderComparison}  
    \resizebox{\textwidth}{!}{%
    \begin{tabular}{c*{9}{c}}
        \toprule
        \multicolumn{1}{c}{Dataset} & \multicolumn{3}{c}{Outdoor scenes} & \multicolumn{3}{c}{Indoor scenes} & \multicolumn{3}{c}{Average on all scenes} \\
        \cmidrule(lr){2-4} \cmidrule(lr){5-7} \cmidrule(lr){8-10}
         \multicolumn{1}{c}{Method \textbar\ Metric} & SSIM $\uparrow$ & PSNR $\uparrow$ & LPIPS $\downarrow$ & SSIM $\uparrow$ & PSNR $\uparrow$ & LPIPS $\downarrow$ & SSIM $\uparrow$ & PSNR $\uparrow$ & LPIPS $\downarrow$ \\
        \midrule
        3DGS~\cite{kerbl20233d} & 0.742 & \cellcolor{red!50}\textbf{25.03} & 0.232 & \cellcolor{orange!50}0.931 & \cellcolor{red!50}\textbf{31.20} & \cellcolor{yellow!50}0.164 & \cellcolor{orange!50}0.837 & \cellcolor{red!50}\textbf{28.12} & \cellcolor{yellow!50}0.198 \\
        2DGS~\cite{huang20242d} & 0.703 & 24.17 & 0.287 & 0.910 & 30.06 & 0.214 & 0.807 & 27.12 & 0.250 \\
        GOF~\cite{yu2024gaussian} & \cellcolor{yellow!50}0.746 & \cellcolor{orange!50}24.81 & \cellcolor{yellow!50}0.208 & 0.917 & \cellcolor{yellow!50}30.40 & 0.189 & \cellcolor{yellow!50}0.832 & \cellcolor{yellow!50}27.61 & 0.199 \\
        PGSR~\cite{chen2024pgsr} & \cellcolor{red!50}\textbf{0.752} & 24.74 & \cellcolor{orange!50}0.203 & \cellcolor{yellow!50}0.929 & 30.16 & \cellcolor{orange!50}0.159 & \cellcolor{red!50}\textbf{0.841} & 27.45 & \cellcolor{orange!50}0.181 \\
        Ours & \cellcolor{orange!50}0.749 & \cellcolor{yellow!50}24.78 & \cellcolor{red!50}\textbf{0.199} & \cellcolor{red!50}\textbf{0.932} & \cellcolor{orange!50}30.66 & \cellcolor{red!50}\textbf{0.151} & \cellcolor{red!50}\textbf{0.841} & \cellcolor{orange!50}27.72 & \cellcolor{red!50}\textbf{0.175} \\
        \bottomrule
    \end{tabular}%
    }
\end{table*}
\begin{figure*}[b]
    \centering
    \includegraphics[width=\textwidth]{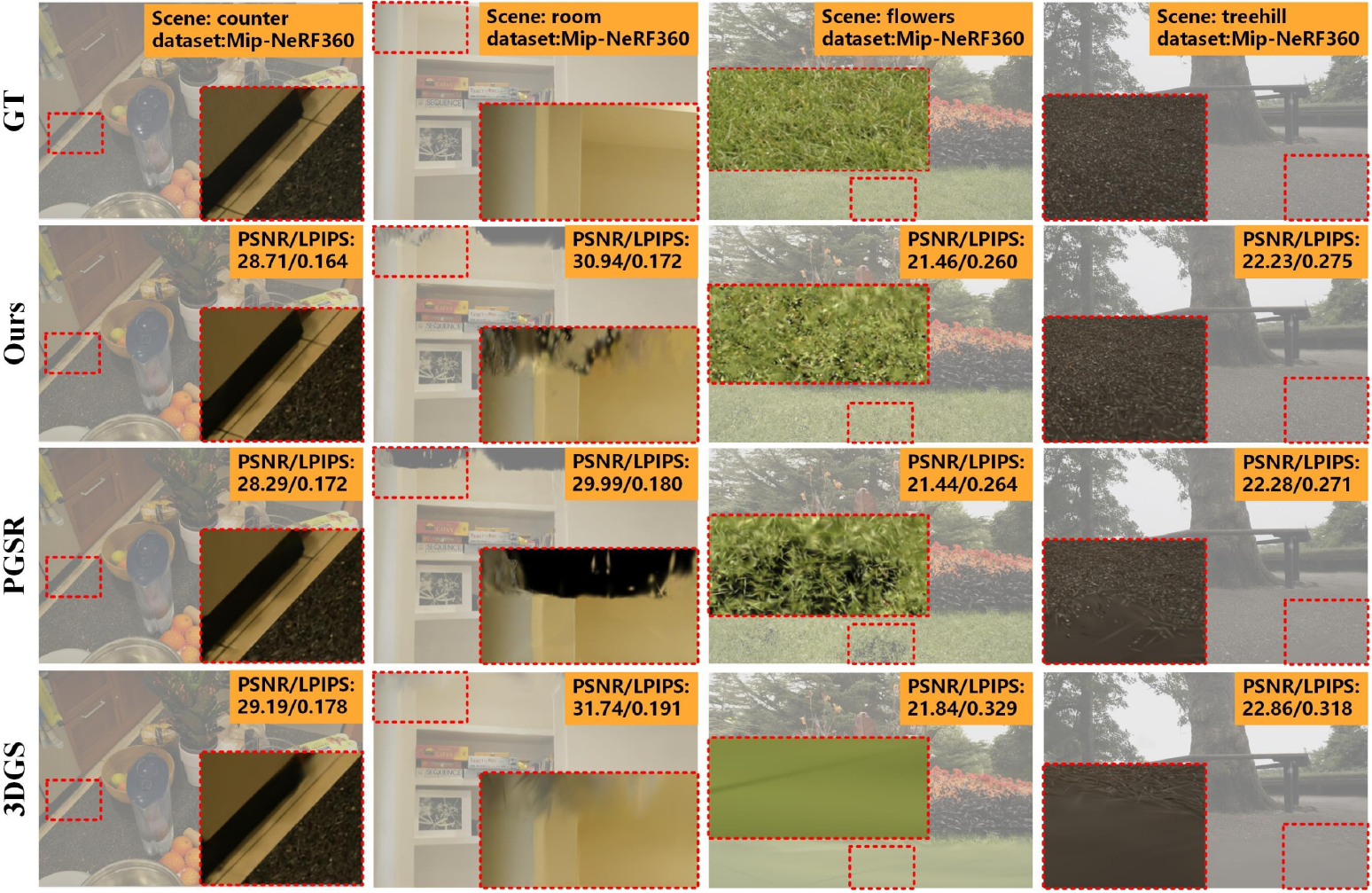} 
    \caption{The figure shows the quantitative analysis of 3DGS, RaDe-GS, PGSR, and the present algorithm in terms of rendering effect, where GT represents the real scene. The rendering effects of four different scenes in the Mip-NeRF360 dataset, including indoor and outdoor, are compared in detail, and the corresponding PSNR and LPIPS metrics of each algorithm are listed. }
    \label{fig:renderComparison}
\end{figure*}
\subsection{Geometry Reconstruction}
The geometric reconstruction performance of the algorithm is analyzed on the DTU and Tanks and Temple datasets, respectively.
Qualitative and quantitative results are shown in Table~\ref{tab:quantitativeDTU}, Table~\ref{tab:quantativeTanks&Temple} and Fig.~\ref{fig:reconstruction}. 

In terms of training efficiency (rightmost column of Table~\ref{tab:quantitativeDTU}), our method requires an average of 0.45 h. This is slightly higher than RaDe-GS (0.13 h), 2DGS (0.15 h), and PGSR (0.28 h). However, it substantially improves geometric accuracy at a modest increase in computation cost. The other methods have CD values of 0.68, 0.76, and 0.53 mm, while our method achieves 0.51 mm. It is also more efficient than the high-accuracy GOF (0.58 h). Ablation studies in the Appendix~\ref{app:Appendix_results_on_dtu} show that the single-view constraint ($\mathcal{L}_{svgeo}$) adds negligible computation (0.28 h $\to$ 0.30 h). The increase in total time mainly comes from the cross-view constraint ($\mathcal{L}_{mvgeo}$), which involves explicit $\mathrm{SE}(3)$ rigid transformations and per-point PCA spectral analysis. These results confirm that our method achieves high-fidelity geometric reconstruction while maintaining computational efficiency.

To evaluate the generalization capability of our method in complex scenes, we conducted group-wise comparisons of F1-Score on the Training subset of the Tanks and Temple dataset. 
The results in Table ~\ref{tab:quantativeTanks&Temple} show that our method achieves an average F1-Score of 0.36, corresponding to improvements of 16.4\%–63.6\% over 2DGS (0.22), GOF (0.31), and RaDe-GS (0.31), and a 5.9\% relative improvement over the second-best method PGSR (0.34), demonstrating its overall advantage in complex outdoor scene reconstruction. 
On the DTU dataset, our method attains a reconstruction accuracy with an average chamfer distance of 0.51 mm, representing a 3.8\% improvement over PGSR (0.53 mm). 
It achieves the best results in 13 out of 15 evaluation scenes and reduces the error by 9.4\% in Scan 110, indicating high robustness and consistency in geometric reconstruction quality.

\subsection{Novel View Synthesis}
\begin{figure}[t]
    \centering
    \includegraphics[width=\columnwidth]{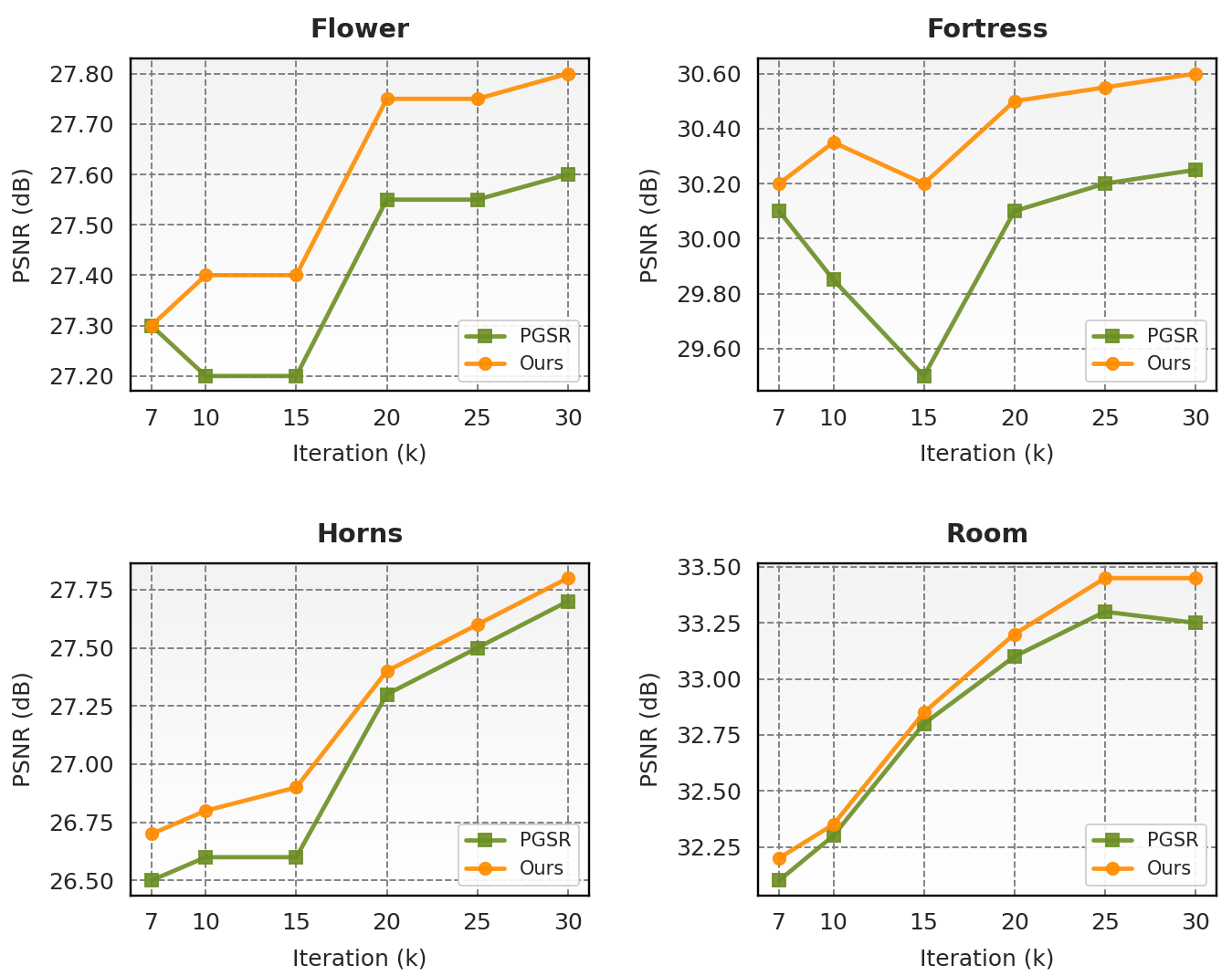} 
    \caption{Quantitative analysis on the LLFF dataset. Based on the comparison of PSNR metric changes during the experiments, it can be seen that our method has better algorithmic performance.}
    \label{fig:LLFF_PSNR}
\end{figure}

\begin{figure*}[t]
    \centering
    \includegraphics[width=\textwidth]{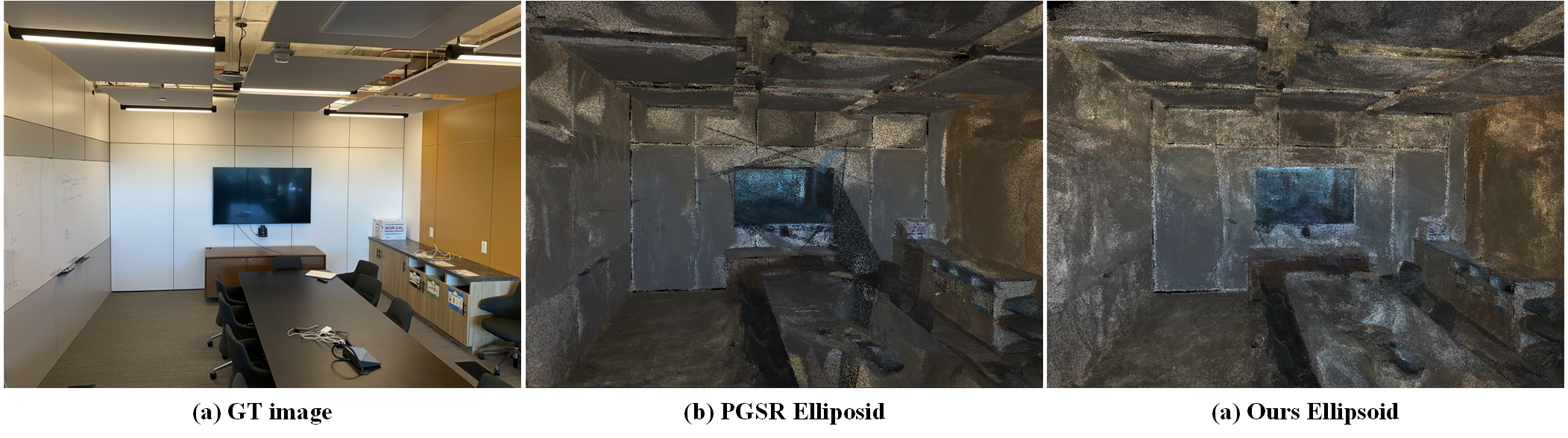} 
    \caption{Comparison of spatial Gaussian ellipsoid distributions. (a) represents the real RGB map; (b) represents the spatial Gaussian ellipsoid distribution of the PGSR method; (c) represents the spatial Gaussian ellipsoid distribution of our method. It can be found that the Gaussian ellipsoid of our method fits the real surface of the object better.}
    \label{fig:ellpisoid}
\end{figure*}
To validate the rendering quality, three (3DGS, PGSR, GSM-GS) better-performing algorithms were selected for quantitative and qualitative analyses on the Mip-NeRF360 dataset, as shown in Table~\ref{tab:renderComparison} and Fig.~\ref{fig:renderComparison}.
The proposed method achieves competitive results across all three key metrics, SSIM, PSNR, and LPIPS, with particularly outstanding LPIPS performance in both indoor and outdoor scenarios. 
For the counter indoor scene, our approach accurately restores line segment details on the floor, closely approximating the real scene. 
In the treehill outdoor scene, it effectively eliminates floor unevenness and distortion artifacts prevalent in other algorithms. 
This advantage stems from our branching geometric constraint strategy, which weights plausible regions and applies differentiated regularisation to texture-rich and texture-deficient areas, thereby capturing geometric details throughout the scene. 
For room and flowers scenes, the introduced cross-view 3D point cloud geometric constraints significantly enhance spatial consistency, particularly in preserving geometric coherence of complex 3D structures, demonstrating the method's competitive rendering capabilities.
More detailed experimental settings and implementation details are provided in Table~\ref{tab:appendix_mipnerf360} of Appendix~\ref{app:Appendix_results_on_mip}.

To demonstrate the superiority of the algorithm proposed in this paper over the benchmark method (PGSR), we conducted extensive experiments on the LLFF dataset~\cite{LLFF_dataset}. 
The PSNR metrics at 7000, 10000, 15000, 20000, 25000, and 30000 iterations are documented, as shown in Fig.~\ref{fig:LLFF_PSNR}. Analysis of the results indicates that our method uniquely incorporates 2D spatial fine-detail capture and 3D spatial point cloud geometry constraints. 
This integration results in a more stable PSNR enhancement trend and a higher final value throughout the iterative process. After 15000 iterations, system densification ceases, and the point cloud count stabilizes.
Our method imposes more effective geometric constraints on the 3D Gaussian point clouds, leading to a closer fit of the Gaussian ellipsoid distribution to the real scene surface, as shown in Fig.~\ref{fig:ellpisoid}. 
The progressively widening PSNR gap relative to PGSR demonstrates improved accuracy in 3D scene reconstruction and validates the enhanced capability of our approach to characterize the scene.

\subsection{Ablation Studies}
\begin{figure}[h]
    \centering
    \includegraphics[width=\columnwidth]{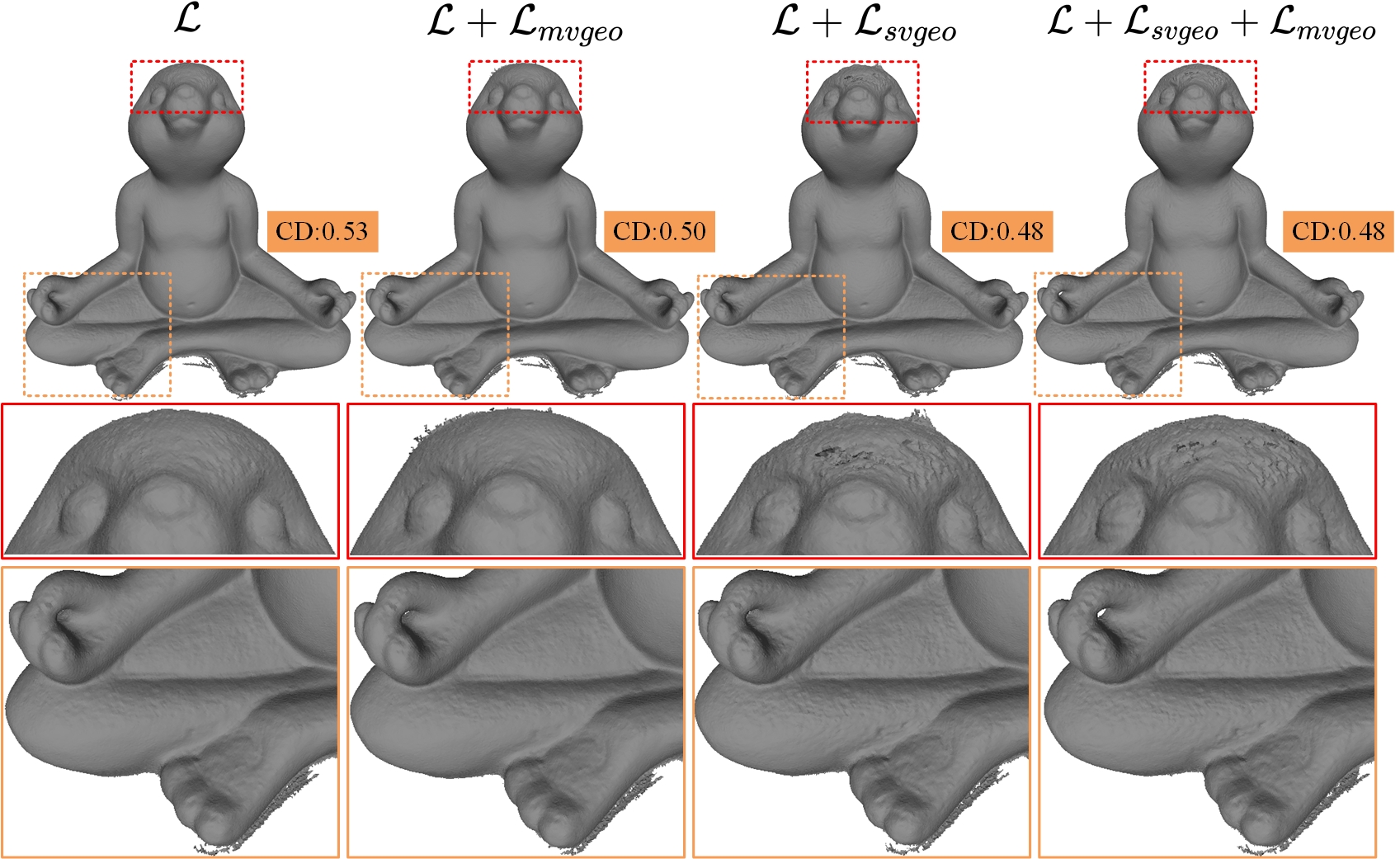}
    \caption{Ablation study on DTU datasets. Observing the details of the reconstructed model, it is found that all of our proposed optimization constraints are able to improve compared to the previous method, with a finer characterization of the surface details.}
    \label{fig:dtuablationfig}
\end{figure}

\begin{figure}[t]
    \centering
    \includegraphics[width=\columnwidth]{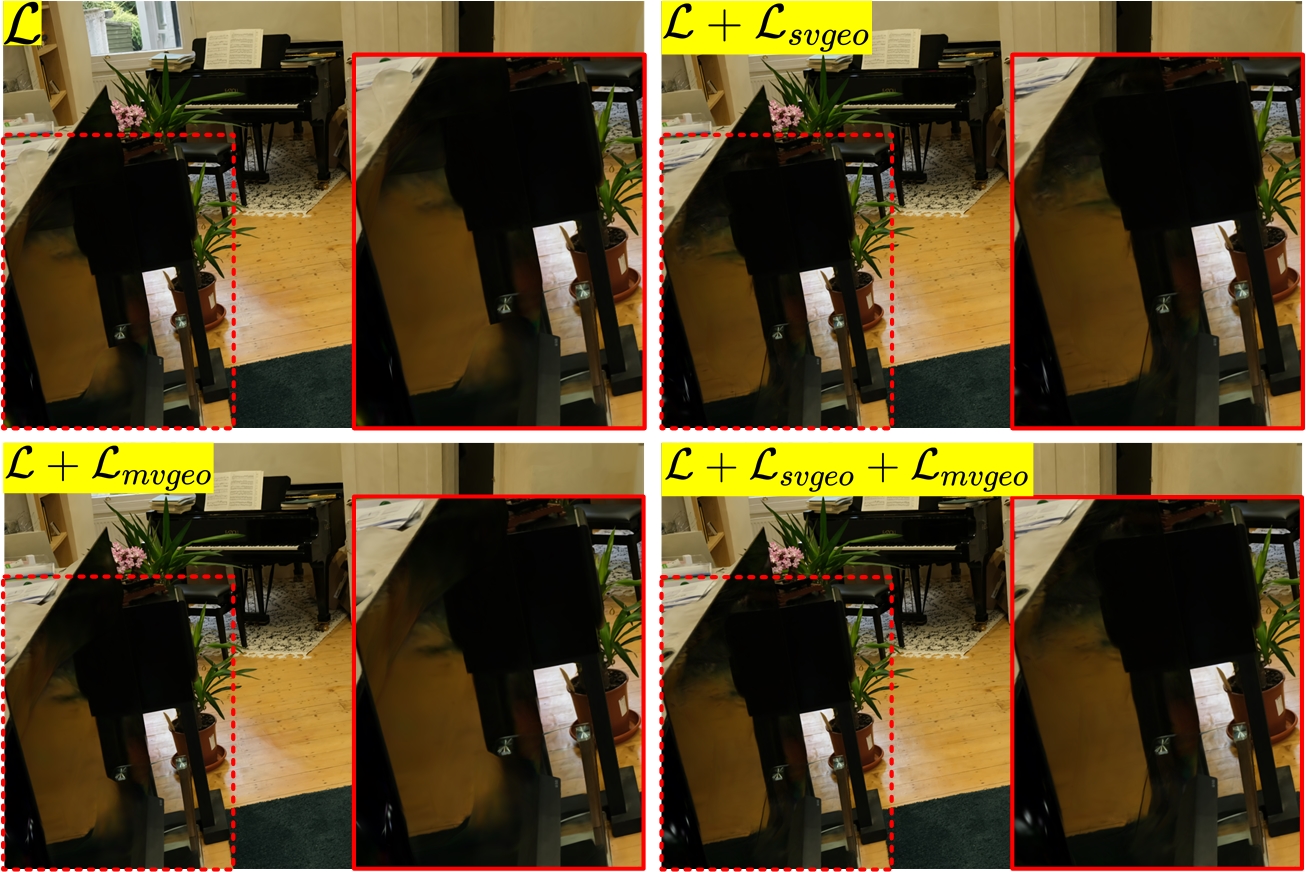}
    \caption{Ablation study on  Mip-NeRF360 datasets. Experiments with novel view synthesis on the  Mip-NeRF360 dataset demonstrate that all of our proposed innovations improve rendering somewhat.}
    \label{fig:mipnerfablationfig}
\end{figure}

To verify the effectiveness of the texture-guided branching-based constraints and surface normal constraints between cross-view point cloud frames proposed in this paper for single-view, we conducted ablation experiments on the DTU and Mip-NeRF360 datasets. 
The results of the qualitative and quantitative analyses are shown in Table~\ref{tab:dtuablationtable}, Table~\ref{tab:mipnerfablationtable}, Fig.~\ref{fig:dtuablationfig}, and Fig.~\ref{fig:mipnerfablationfig}. Among them, $\mathcal{L}$ denotes the benchmark constraints of the baseline method~\cite{chen2024pgsr}, which includes image photometric consistency, single-view geometric constraints, and multi-view geometric constraints; $\mathcal{L}_{svgeo}$ and $\mathcal{L}_{mvgeo}$ represent the single-view and multi-view geometric constraints, respectively, innovatively designed in this paper's optimization modules. 
Therefore, $(\mathcal{L}+\mathcal{L}_{svgeo})$ and $(\mathcal{L}+\mathcal{L}_{mvgeo})$ denote the single-view and multi-view geometric constraints schemes, respectively, in the baseline method optimized with the innovative constraints proposed in this paper.

\begin{table}[h]
    \centering
    \caption{Ablation study of our model on reconstruction accuracy across the DTU datasets.}  
    \resizebox{\columnwidth}{!}{ 
    \begin{tabular}{cccccc}
        \toprule
        $\quad \mathcal{L} \quad$ & $ \quad \mathcal{L}_{svgeo} \quad$ & $\quad \mathcal{L}_{mvgeo} \quad  $ & CD$\downarrow$ \\
        \midrule
            \checkmark & \texttimes & \texttimes & 0.53  \\
            \checkmark & \checkmark & \texttimes & 0.50  \\
            \checkmark & \texttimes & \checkmark & \textbf{0.48}  \\
            \checkmark & \checkmark & \checkmark & \textbf{0.48}  \\
        \bottomrule
    \end{tabular}
    }
    \label{tab:dtuablationtable}
\end{table}

Comparison of the experimental data shows that, compared to the benchmark method $\mathcal{L}$, introducing either the single-view optimization constraints($\mathcal{L}+\mathcal{L}_{svgeo}$) or the multi-view optimization constraints($\mathcal{L}+\mathcal{L}_{mvgeo}$) significantly improves both 3D reconstruction accuracy and new view synthesis quality. Furthermore, the synergistic integration of the two constraints ($\mathcal{L}+\mathcal{L}_{svgeo}+\mathcal{L}_{mvgeo}$) drives the system to achieve better performance. 
Specifically, the single-view texture-guided branching constraint significantly enhances robustness and generalization ability to dynamic scene changes by modeling multi-scale scene detail features. 

\begin{table}[t]
    \centering
    \caption{Ablation study of our model on novel view synthesis across the  Mip-NeRF360 datasets.}  
    \resizebox{\columnwidth}{!}{ 
    \begin{tabular}{cccccc}
        \toprule
        $\mathcal{L}$ & $ \mathcal{L}_{svgeo}$ & $\mathcal{L}_{mvgeo} $ &SSIM$\uparrow$ & PSNR$\uparrow$ & LPIPS $\downarrow$ \\
        \midrule
            \checkmark & \texttimes & \texttimes & 0.928 & 30.22 & 0.179   \\
            \checkmark & \checkmark & \texttimes & 0.930 & 30.77 & \textbf{0.172}   \\
            \checkmark & \texttimes & \checkmark & 0.929 & 30.35 & 0.179   \\
            \checkmark & \checkmark & \checkmark & \textbf{0.930} & \textbf{30.80} & \textbf{0.172}   \\
        \bottomrule
    \end{tabular}
    }
    \label{tab:mipnerfablationtable}
\end{table}


Concurrently, the cross-view point cloud inter-frame surface normal constraint effectively eliminates geometric artifacts in 3D space by reinforcing multi-view geometric consistency, thereby optimizing reconstruction rendering. 
Taken together, each constraint mechanism proposed in this paper independently improves the baseline model's performance, with all modules contributing significantly to enhancing the final reconstruction and rendering results.
More detailed ablation results on scene reconstruction are provided in Appendix~\ref{app:Appendix_results_on_dtu}.

\section{Conclusion}
\label{app:conclusion}
This paper proposes GSM-GS, which achieves region partitioning and adaptive constraints through single-view gradients and adaptive weighting.
Furthermore, it introduces a geometry-guided cross-view point cloud association mechanism to effectively enforce multi-view consistency within the 3D space.
The core innovation of this work lies in the proposal of region-specific customized constraint strategies and the construction of a multi-view consistency enhancement mechanism based on 3D spatial characteristics, which effectively suppresses geometric artifacts.
Extensive experiments on public datasets featuring challenging scenarios, such as texture-less regions, occlusions, and specular highlights, demonstrate the superiority and robustness of the proposed method in both surface reconstruction and novel view synthesis tasks.

Nevertheless, the method still faces challenges when dealing with transparent materials, highly reflective surfaces, and intricate thin-walled structures due to the inherent ambiguity between geometry and appearance induced by complex lighting effects.
Future work will focus on exploring more advanced scene representations and optimization strategies to further enhance the model's robustness and generalization capabilities in complex scenes.

\bibliographystyle{IEEEtran}
\bibliography{IEEEfull}

\clearpage
\section*{\bf APPENDIX}
\appendices
This section is structured as follows: Appendix~\ref{app:symbol_definition} provides a systematic definition of the key notations employed throughout this paper; Appendix~\ref{app:single_view_and_multi} presents an in-depth sensitivity analysis of the core hyperparameters; and Appendix~\ref{app:additional_results} offers supplementary qualitative and quantitative evaluations across multiple datasets to further validate the proposed method.

\section{ Symbol Definition }
\label{app:symbol_definition}
In this section, we provide a detailed explanation of the key symbols involved in the methodology, as shown in Table~\ref{tab:symbol_definition}.

\begin{table}[h]
    \centering
    \caption{Summary of Mathematical Notations and Symbols}
    \label{tab:symbol_definition}
    \renewcommand{\arraystretch}{1.3} 
    \resizebox{\columnwidth}{!}{%
    \begin{tabular}{cc}
        \toprule
        \textbf{Symbol} & \textbf{Definition} \\
        \midrule
        $\Omega$ & Image domain \\
        $\mathcal{H}$ &  High-weight trust regions \\
        $\theta$ & Trust regions threshold \\
        $\mathcal{R}, \mathcal{B}$ & Texture-rich and texture-less regions \\
        $\tau$ & Gradient Threshold \\
        $\mathcal{M}$ & Geometric validity mask \\
        $\epsilon_d$ & Depth Threshold \\
        $\gamma$ & Adaptive dynamic threshold \\
        $\mathcal{Q}, \mathcal{Q}_s$ & Candidate set and high-confidence regions \\
        $S$ & Sampling Rate \\
        $\text{rank}(\cdot)$ & Descending sorting operator \\
        $\mathbf{\Pi}_3(\cdot)$ & Projection operator (Homogeneous to Euclidean) \\
        \bottomrule
    \end{tabular}%
    }
\end{table}

\section{Single-View and Multi-View Parameter Settings}
\label{app:single_view_and_multi}
This section investigates the sensitivity of key parameters under both single-view and multi-view conditions. Specifically, we evaluate the trust threshold $\theta$ for distinguishing reliable from unreliable regions, the gradient-based threshold $\tau$ for segregating texture-rich and texture-less areas, the depth threshold $\epsilon_d$, the adaptive dynamic threshold $\gamma$, and sampling rate $S$.
\begin{figure*}[hpb!]
    \centering
    \includegraphics[width=\textwidth]{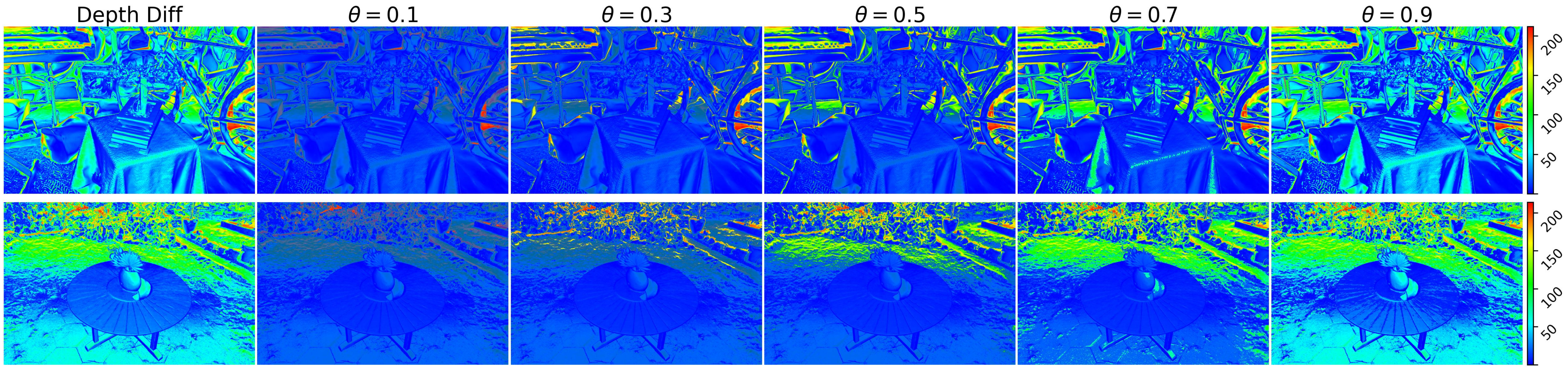}
    \caption{Sensitivity analysis of the confidence threshold $\theta$. The optimal value ($\theta = 0.8$) aligns high-weight regions (blue mask) with low-error areas in the depth difference map, providing reliable region selection and balancing accuracy with robustness.}
    \label{fig:appendix_theta_threshold}
\end{figure*}
\begin{figure*}[t]
    \centering
    \includegraphics[width=\textwidth]{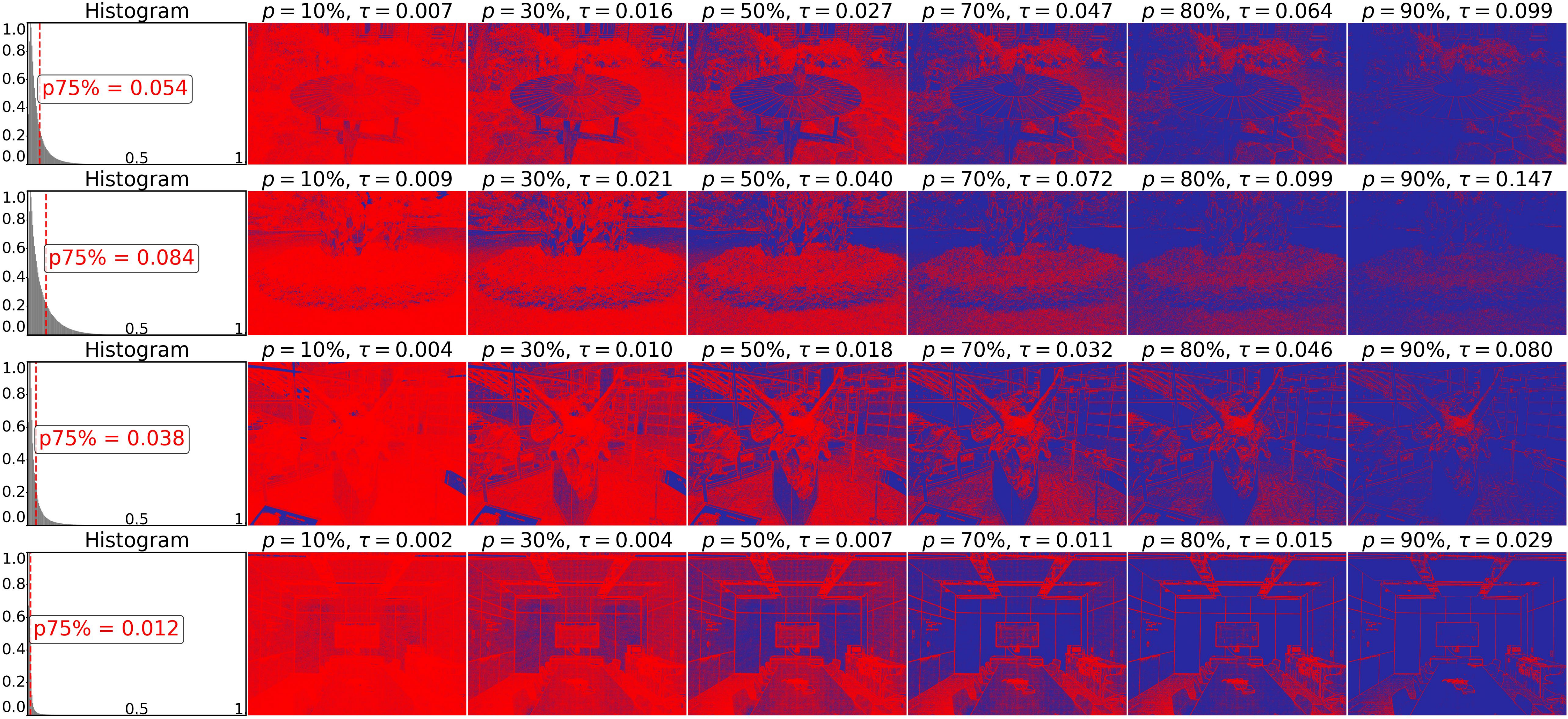}
    \caption{Sensitivity analysis of texture segmentation quantiles. Across diverse scenes, $p \in [50\%, 80\%]$ provides the most stable partitioning, while other ranges cause noise over-segmentation or edge loss. Thus, $p=75\%$ is selected as the optimal setting for robust texture-less region separation.}
    \label{fig:ablation_tau_threshold}
\end{figure*}

\subsection{Single-view Parameter Settings}
\label{app:Appendix_single_view_parameter}
To evaluate the efficacy of the exploration threshold $\theta$ in identifying reliable regions, the selected high-weight regions are superimposed onto the depth difference map using a semi-transparent blue mask(as shown in Fig.~\ref{fig:appendix_theta_threshold}). 
In these maps, blue/cyan represents low depth error (high confidence), while yellow/red indicates high error (low confidence).
An optimal threshold should ensure that the high-weight mask predominantly covers blue/cyan areas; coverage of yellow/red regions implies misjudgment due to an overly loose threshold.
Observations reveal that for $\theta < 0.5$, the mask erroneously extends into significant yellow/red high-error regions, indicating insufficient separation.
Conversely, within the $0.7 - 0.9$ range, the boundaries become distinct and align well with low-error areas. Consequently, we selected $\theta=0.8$ as the default setting, where the high-weight regions exhibit the best spatial consistency with low-error areas, achieving an optimal balance between accuracy and robustness.

To evaluate the sensitivity of the segmentation performance to the threshold $\tau$ (distinguishing texture-rich from texture-less regions), we systematically analyze gradient magnitude distributions and corresponding quantile thresholds across multiple scenes Fig.~\ref{fig:ablation_tau_threshold}. The leftmost column displays gradient histograms, while subsequent columns illustrate segmentation results derived from thresholds at quantiles ranging from $p=10\%$ to $p=90\%$. Experimental results indicate that low quantiles ($10–30\%$) yield excessively low thresholds, causing widespread misclassification of weak gradient regions as textured areas. Conversely, high quantiles ($90\%$) result in overly aggressive thresholds, leading to the significant loss of structural edges. Only the intermediate range ($50–80\%$) demonstrates stable segmentation consistency across diverse scenes, where the proportion of high-texture regions, denoted as $h_f$, decreases smoothly and monotonically with $p$. Consequently, the optimal range for $\tau$ is constrained to $p \in [50\%, 80\%]$. In this study, we select the threshold corresponding to $p=75\%$, as it resides within this stable interval and ensures robust texture partitioning.

\subsection{Multi-view Parameter Settings}
\label{app:Appendix_multi_view_parameter}
Following the PGSR framework\cite{chen2024pgsr}, we set the depth threshold $\epsilon_d$ in mask $\mathcal{M}_d$ to $0.1m$. This enforces a physical lower bound $z_i \geq \epsilon_d$, effectively filtering near-field high-frequency noise and low-SNR artifacts to enhance reconstruction robustness.

We introduce an adaptive threshold $\gamma$, set to $30\%$ of the global average weight $W_{avg}$, to enforce a minimum quality baseline for the candidate set $\mathcal{Q}$. By acting as a preliminary filter, $\gamma$ excludes low-confidence points to ensure $W_{avg}(i, j) \geq \gamma$, thereby guaranteeing a valid geometric lower bound. This configuration maintains a sufficient candidate pool ($|\mathcal{Q}| \gg S$) while securing the robustness of the sampling strategy across diverse scenarios.
\begin{figure}[h]
    \centering
    \includegraphics[width=\columnwidth]{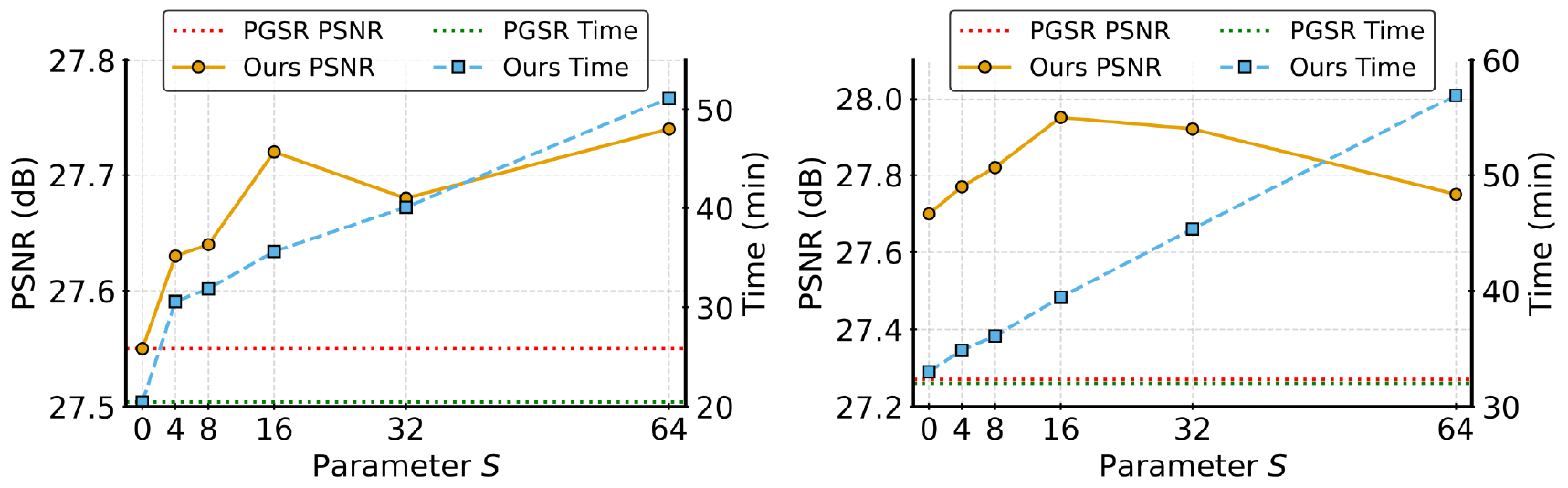}
    \caption{Sensitivity analysis of sampling parameter $S$ regarding PSNR and total computation time. The yellow solid lines represent our PSNR (dB) and the blue dashed lines represent total training time (min). The horizontal red and green dotted lines denote the baseline performance of PGSR. Peak accuracy is achieved at $S=16$, after which the accuracy gains diminish while computational costs continue to rise linearly.}
    \label{fig:appendix_s_threshold}
\end{figure}


A sensitivity analysis was conducted on the key parameter $S$ of the weight-guided geometric sampling strategy to balance reconstruction accuracy against computational efficiency (as shown in Fig.~\ref{fig:appendix_s_threshold} and Table~\ref{tab:appendix_s_table}). 
The reconstruction accuracy exhibited a trend of rapid initial improvement followed by saturation and a marginal decline as $S$ increased; specifically, the PSNR peaked at 27.95 dB when $S=16$. 
Compared to the low sampling rate of $S=4$ (27.77 dB), this 0.18 dB improvement validates the efficacy of moderate sampling in capturing geometric details. 
However, when $S > 16$, the accuracy gains diminished or even slightly regressed (27.75 dB at $S=64$).
This is primarily because an excessive sampling count introduces points with low weights and insufficient geometric confidence, allowing geometric noise to interfere with the overall optimization process.
Meanwhile, the total computation time increased linearly and substantially, from 39.44 min at $S=16$ to 56.96 min at $S=64$. Consequently, $S=16$ achieves optimal accuracy while maintaining acceptable computational overhead, representing the ideal trade-off between precision and efficiency; it is thus selected as the default sampling parameter.
\begin{table}[t!]
\centering
\caption{Quantitative impact of sampling parameter $S$ on reconstruction performance and time consumption.}
\label{tab:appendix_s_table}
\resizebox{\columnwidth}{!}{%
\begin{tabular}{cc|cccccc}
\toprule
\multirow{2}{*}{Scenes} & \multirow{2}{*}{Metric} & \multicolumn{6}{c}{S} \\
\cmidrule(lr){3-8}
 & & 0 & 4 & 8 & \textbf{16} & 32 & 64 \\
\midrule
\multirow{2}{*}{flower} 
 & PSNR $\uparrow$ & 27.55 & 27.63 & 27.64 & \textbf{27.72} & 27.68 & 27.74 \\
 & All Time $\downarrow$ & 20.45 & 30.85 & 31.86 & \textbf{35.63} & 40.10 & 51.15 \\
\midrule
\multirow{2}{*}{trex} 
 & PSNR $\uparrow$ & 27.70 & 27.77 & 27.82 & \textbf{27.95} & 27.92 & 27.75 \\
 & All Time $\downarrow$ & 33.00 & 34.86 & 36.10 & \textbf{39.44} & 45.35 & 56.96 \\
\bottomrule
\end{tabular}%
}
\end{table}


\section{Additional Results}
\label{app:additional_results}
\subsection{Results on Mip-NeRF360} 
\label{app:Appendix_results_on_mip}
Table~\ref{tab:appendix_mipnerf360} presents a scene-wise quantitative comparison on the Mip-NeRF360 dataset, demonstrating our method's robustness against state-of-the-art baselines. Notably, we outperform the strong baseline PGSR in LPIPS across 8 of 9 scenes and achieve superior PSNR in 7 of 9 scenes. These results validate that by imposing strict geometric constraints, our approach effectively mitigates visual artifacts while maintaining competitive rendering fidelity in complex environments.

\subsection{Results on DTU Dataset}
\label{app:Appendix_results_on_dtu}
Comprehensive ablation studies on the DTU dataset (Table~\ref{tab:appendix_dtu}) corroborate the efficacy of our method. Results indicate that incorporating $\mathcal{L}_{svgeo}$ or $\mathcal{L}_{mvgeo}$ individually enhances reconstruction metrics. Their joint application yields performance comparable to individual constraints due to performance saturation, yet it consistently outperforms the baseline. Although these constraints incur a moderate increase in training time, the computational overhead remains manageable. Thus, the proposed method offers a favorable balance between reconstruction fidelity and efficiency.

\begin{table*}[t!]
    \centering
    \caption{Quantitative comparison on the Mip-NeRF 360 dataset. We report SSIM, PSNR, and LPIPS metrics across different scenes. "Red", "Orange", and "Yellow" indicate the best, second-best, and third-best results, respectively.}
    \label{tab:appendix_mipnerf360}
    
    \begin{minipage}{\textwidth}
        \resizebox{\textwidth}{!}{%
            \begin{tabular}{c|c|ccccccccc}
            \toprule
            Method & Metric & bicycle & bonsai & counter & flowers & garden & kitchen & room & stump & treehill \\
            \midrule
            
            \multirow{3}{*}{3DGS~\cite{kerbl20233d}} 
             & SSIM $\uparrow$ & \cellcolor{yellow!50}0.779 & \cellcolor{red!50}\textbf{0.947} & \cellcolor{orange!50}0.915 & 0.622 & \cellcolor{orange!50}0.874 & \cellcolor{orange!50}0.933 & \cellcolor{orange!50}0.928 & \cellcolor{yellow!50}0.783 & \cellcolor{orange!50}0.653 \\
             & PSNR $\uparrow$ & \cellcolor{red!50}\textbf{25.65} & \cellcolor{red!50}\textbf{32.37} & \cellcolor{red!50}\textbf{29.19} & \cellcolor{red!50}\textbf{21.84} & \cellcolor{red!50}\textbf{27.82} & \cellcolor{red!50}\textbf{31.51} & \cellcolor{red!50}\textbf{31.74} & \cellcolor{yellow!50}26.96 & \cellcolor{red!50}\textbf{22.86} \\
             & LPIPS $\downarrow$ & 0.203 & \cellcolor{yellow!50}0.175 & \cellcolor{yellow!50}0.178 & 0.329 & \cellcolor{orange!50}0.103 & \cellcolor{orange!50}0.113 & \cellcolor{yellow!50}0.191 & 0.208 & 0.318 \\
            \midrule
            
            \multirow{3}{*}{2DGS~\cite{huang20242d}} 
             & SSIM $\uparrow$ & 0.731 & 0.929 & 0.890 & 0.572 & 0.838 & 0.914 & 0.905 & 0.756 & 0.618 \\
             & PSNR $\uparrow$ & 24.70 & 31.18 & 28.07 & 21.07 & 26.57 & 30.15 & \cellcolor{yellow!50}30.83 & 26.14 & \cellcolor{yellow!50}22.39 \\
             & LPIPS $\downarrow$ & 0.272 & 0.229 & 0.233 & 0.377 & 0.149 & 0.148 & 0.245 & 0.260 & 0.377 \\
            \midrule
            
            \multirow{3}{*}{GOF~\cite{yu2024gaussian}} 
             & SSIM $\uparrow$ & \cellcolor{orange!50}0.787 & \cellcolor{yellow!50}0.937 & 0.902 & \cellcolor{red!50}\textbf{0.638} & 0.868 & 0.916 & 0.913 & \cellcolor{yellow!50}0.794 & 0.643 \\
             & PSNR $\uparrow$ & 25.48 & 31.57 & \cellcolor{yellow!50}28.69 & \cellcolor{orange!50}21.66 & 27.42 & 30.68 & 30.65 & \cellcolor{orange!50}26.98 & \cellcolor{orange!50}22.49 \\
             & LPIPS $\downarrow$ & \cellcolor{orange!50}0.180 & 0.198 & 0.203 & \cellcolor{yellow!50}0.280 & \cellcolor{yellow!50}0.107 & \cellcolor{yellow!50}0.137 & 0.218 & \cellcolor{yellow!50}0.196 & \cellcolor{yellow!50}0.278 \\
            \midrule
            
            \multirow{3}{*}{PGSR~\cite{chen2024pgsr}} 
             & SSIM $\uparrow$ & \cellcolor{red!50}\textbf{0.793} & \cellcolor{orange!50}0.945 & \cellcolor{yellow!50}0.914 & \cellcolor{orange!50}0.636 & \cellcolor{yellow!50}0.872 & \cellcolor{yellow!50}0.932 & \cellcolor{yellow!50}0.926 & \cellcolor{orange!50}0.797 & \cellcolor{red!50}\textbf{0.661} \\
             & PSNR $\uparrow$ & \cellcolor{orange!50}25.64 & \cellcolor{yellow!50}31.59 & 28.29 & 21.44 & \cellcolor{yellow!50}27.43 & \cellcolor{yellow!50}30.76 & 29.99 & 26.89 & 22.28 \\
             & LPIPS $\downarrow$ & \cellcolor{yellow!50}0.186 & \cellcolor{orange!50}0.169 & \cellcolor{orange!50}0.172 & \cellcolor{orange!50}0.264 & \cellcolor{orange!50}0.103 & \cellcolor{orange!50}0.113 & \cellcolor{orange!50}0.180 & \cellcolor{orange!50}0.193 & \cellcolor{red!50}\textbf{0.271} \\
            \midrule
            
            \multirow{3}{*}{Ours}
             & SSIM $\uparrow$ & \cellcolor{red!50}\textbf{0.793} & \cellcolor{red!50}\textbf{0.947} & \cellcolor{red!50}\textbf{0.917} & \cellcolor{yellow!50}0.633 & \cellcolor{red!50}\textbf{0.875} & \cellcolor{red!50}\textbf{0.934} & \cellcolor{red!50}\textbf{0.930} & \cellcolor{red!50}\textbf{0.798} & \cellcolor{yellow!50}0.648 \\
             & PSNR $\uparrow$ & \cellcolor{yellow!50}25.61 & \cellcolor{orange!50}31.97 & \cellcolor{orange!50}28.71 & \cellcolor{yellow!50}21.46 & \cellcolor{orange!50}27.59 & \cellcolor{orange!50}31.03 & \cellcolor{orange!50}30.94 & \cellcolor{red!50}\textbf{27.00} & 22.23 \\
             & LPIPS $\downarrow$ & \cellcolor{red!50}\textbf{0.178} & \cellcolor{red!50}\textbf{0.160} & \cellcolor{red!50}\textbf{0.164} & \cellcolor{red!50}\textbf{0.260} & \cellcolor{red!50}\textbf{0.098} & \cellcolor{red!50}\textbf{0.109} & \cellcolor{red!50}\textbf{0.172} & \cellcolor{red!50}\textbf{0.186} & \cellcolor{orange!50}0.275 \\
            \bottomrule
            \end{tabular}%
        }
    \end{minipage}
\end{table*}
\begin{table*}[t!]
\centering
\caption{Supplementary ablation studies were conducted using the DTU dataset. Best results are highlighted in red.}
\label{tab:appendix_dtu}
\resizebox{\textwidth}{!}{%
\begin{tabular}{c|ccccccccccccccc|cc}
\toprule
\textbf{Method} & \textbf{24} & \textbf{37} & \textbf{40} & \textbf{55} & \textbf{63} & \textbf{65} & \textbf{69} & \textbf{83} & \textbf{97} & \textbf{105} & \textbf{106} & \textbf{110} & \textbf{114} & \textbf{118} & \textbf{122} & \textbf{Mean} & \textbf{Mean Time} \\
\midrule
$\mathcal{L}$ & 0.37 & 0.55 & 0.42 & 0.35 & 0.78 & 0.58 & 0.49 & 1.08 & 0.64 & 0.59 & 0.48 & 0.53 & 0.30 & 0.37 & 0.35 & 0.53 & \cellcolor{red!50}\textbf{0.28h} \\
$+\mathcal{L}_{mvgeo}$ & 0.37 & \cellcolor{red!50}\textbf{0.55} & 0.39 & 0.34 & 0.78 & 0.58 & 0.49 & 1.06 & 0.64 & 0.58 & 0.47 & 0.50 & 0.30 & 0.37 & 0.34 & 0.52 & 0.45h \\
$+\mathcal{L}_{svgeo}$ & 0.36 & 0.56 & \cellcolor{red!50}\textbf{0.36} & 0.34 & 0.77 & 0.58 & 0.49 & 1.05 & \cellcolor{red!50}\textbf{0.63} & 0.58 & 0.47 & 0.48 & 0.30 & 0.36 & 0.33 & 0.51 & 0.30h \\
$\mathcal{L}+\mathcal{L}_{svgeo}+\mathcal{L}_{mvgeo}$ & \cellcolor{red!50}\textbf{0.36} & 0.56 & 0.37 & \cellcolor{red!50}\textbf{0.34} & \cellcolor{red!50}\textbf{0.77} & \cellcolor{red!50}\textbf{0.57} & \cellcolor{red!50}\textbf{0.49} & \cellcolor{red!50}\textbf{1.03} & 0.64 & \cellcolor{red!50}\textbf{0.58} & \cellcolor{red!50}\textbf{0.47} & \cellcolor{red!50}\textbf{0.48} & \cellcolor{red!50}\textbf{0.30} & \cellcolor{red!50}\textbf{0.36} & \cellcolor{red!50}\textbf{0.33} & \cellcolor{red!50}\textbf{0.51} & 0.45h \\
\bottomrule
\end{tabular}%
}
\end{table*}

\end{document}